%% file: main.tex
\documentclass{article} 
\usepackage[preprint]{timeseries_workshop}

\usepackage{pdflscape}
\usepackage[utf8]{inputenc} 
\usepackage[T1]{fontenc}    
\usepackage{hyperref}       
\usepackage{url}            
\usepackage{booktabs}       
\usepackage{amsfonts}       
\usepackage{nicefrac}       
\usepackage{microtype}      
\usepackage{xcolor}         
\usepackage{booktabs}   
\usepackage{multirow}   
\usepackage{ltablex}    
\keepXColumns  

\newcommand{\thead}[1]{\texttt{\begin{tabular}{@{}c@{}}#1\end{tabular}}}

\usepackage{adjustbox}

\usepackage{booktabs}
\usepackage{multirow}
\usepackage{longtable}
\usepackage[table]{xcolor}

\usepackage{multirow}
\usepackage{amsmath}
\usepackage{amssymb}
\usepackage[capitalize]{cleveref}
\usepackage{colortbl}
\usepackage[inline,shortlabels]{enumitem}
\usepackage{etoolbox}
\usepackage{array}
\usepackage{booktabs}      
\usepackage{xcolor}        
\usepackage{pifont}        
\usepackage{soul}

\usepackage{graphicx}
\usepackage{subcaption}



\definecolor{edf1}{rgb}{0, 0.102, 0.439}
\definecolor{edf2}{rgb}{0, 0.357, 0.733}
\definecolor{edf3}{rgb}{1, 0.627, 0.184}
\definecolor{edf4}{rgb}{0.996, 0.345, 0.082}
\definecolor{edf5}{RGB}{196, 214, 0}
\definecolor{edf6}{RGB}{80, 158, 47}

\colorlet{maincolor}{edf1}

\definecolor{bgcolor}{HTML}{ECF1FC}
\definecolor{fgcolor}{HTML}{222244}

\definecolor{colframe}{gray}{0.2}
\colorlet{colback}{bgcolor}

\usepackage[most]{tcolorbox}
\usepackage[title]{appendix}

\title{On the Role of Reversible Instance Normalization}
\date{October 2025}
\author{
Gaspard Berthelier \\
EDF R\&D \& Inria Sophia-Antipolis \\
\texttt{\footnotesize gaspard.berthelier@edf.fr}
\And
Tahar Nabil \\
EDF R\&D \\
\texttt{\footnotesize tahar.nabil@edf.fr}
\And
Etienne Le Naour \\
EDF R\&D \\
\texttt{\footnotesize etienne.le-naour@edf.fr}
\AND
Richard Niamke\\
EDF R\&D \\
\texttt{\footnotesize richard.niamke@edf.fr}
\And
Samir Perlaza\\
Inria Sophia-Antipolis\\
\texttt{\footnotesize samir.perlaza@inria.fr}
\And
Giovanni Neglia\\
Inria Sophia-Antipolis\\
\texttt{\footnotesize giovanni.neglia@inria.fr}
}

\begin{document}

\maketitle

\input{00--abstract}
\input{01--intro}
\input{02--normalization-methods}
\input{03--experiments}
\input{04--conclusion}

\bibliography{biblio}
\bibliographystyle{neurips_2024}

\newpage

\input{appendix.tex}

\end{document}

%% file: 00--abstract.tex
\begin{abstract}
Data normalization is a crucial component of deep learning models, yet its role in time series forecasting remains insufficiently understood. In this paper, we identify three central challenges for normalization in time series forecasting: temporal input distribution shift, spatial input distribution shift, and conditional output distribution shift. In this context, we revisit the widely used Reversible Instance Normalization (RevIN), by showing through ablation studies that several of its components are redundant or even detrimental. Based on these observations, we draw new perspectives to improve RevIN's robustness and generalization.
\end{abstract}

%% file: 01--intro.tex
\section{Introduction} \label{intro}

Data normalization is a fundamental preprocessing step, long recognized as essential for the stable and efficient training of neural networks. Its significance was established in early studies on network optimization \citep{DBLP:series/lncs/LeCunBOM12}. 
Normalization is particularly critical in time series forecasting, a domain in which deep learning models have achieved state-of-the-art performance \citep{patchtst}. Intrinsic properties of time series, such as non-stationarity, trends, and seasonality, introduce challenges that standard normalization methods fail to address \citep{adaptivenorm}. 

Three major aspects must be considered when designing a normalization strategy for time series forecasting (formalized in \cref{apx:shifts}):

\begin{enumerate}[(i)]
    \item \textbf{Temporal distribution shift.} When training a neural forecaster over a training period, the input data distribution in a future test period may differ substantially. For instance, electricity consumption or road traffic often increase over time (see \cref{fig:a}).

    \item \textbf{Spatial distribution shift.} Neural forecasters are typically trained on multiple time series and must generalize to unseen time series at inference. Even when series describe similar phenomena (e.g., solar power generation at different locations), their overall distributions may differ, for example, due to scale or level shifts (see \cref{fig:b}).

    \item \textbf{Conditional distribution shift.} A neural forecaster maps past \textit{look-back} windows to future \textit{horizon} windows. The conditional distribution of the horizon given the look-back may vary both in space and time. This shift remains particularly challenging to handle (see \cref{fig:c}).
\end{enumerate}

\input{plots/distrib-shift}


In deep learning, a common approach is to apply \textit{standard normalization} to all data points, using a single global mean $\mu$ and standard deviation $\sigma$ computed empirically over the training data. Each data point is processed by subtracting the mean and dividing by the standard deviation. However, this global normalization fails to address the challenges of the three distribution shifts
described above 
\citep{adaptivenorm}. In time series forecasting, a more sophisticated and widely adopted normalization strategy is \textit{Reversible Instance Normalization} (RevIN) \citep{revin}.

The RevIN method has been adopted in numerous recent forecasting studies \citep{patchtst,fits,tirex,HUANG2025114036,revin}. Its authors claim that RevIN mitigates the distribution shift problem in time series. In this work, we challenge this claim. 
Our contributions are summarized as follows:
\begin{itemize}
    \item We review normalization techniques for neural forecasters and identify the key challenges they face in time series applications.
    \item We conduct extensive ablation studies on RevIN using standard forecasting benchmarks and show which components are truly necessary.
    \item We discuss the effects of instance normalization on data distributions and highlight the limitations of RevIN.
\end{itemize}

\section{Setting}


Time series forecasting consists in finding an optimal parameterized model $f_\theta$ of the form:
\begin{equation}
\label{eq:definition}
f_\theta:\;
\begin{array}{@{}c@{~}c@{~}l@{}}
  \mathbb{R}^{L} & \longrightarrow & \mathbb{R}^{H}, \\
x & \longmapsto & \hat{y} \approx y,
\end{array}
\qquad
\begin{aligned}
L &:\ \text{look-back window size},\\
H &:\ \text{horizon size}.
\end{aligned}
\end{equation}
where $\theta$ corresponds to the parameters of the model (e.g. a neural network), fitted on a training dataset $\mathcal{D}$ of $N$ pairs of windows $(x, y)$, with $x$ a past look-back window and $y$ a horizon to forecast. Naturally, the goal is to have predictions $\hat{y}$ as close as possible to the ground-truth horizons $y$. We say a model \textit{generalizes} well when this is the case for unseen windows. 

\cref{eq:definition} corresponds more precisely to the \textit{auto-regressive} and \textit{univariate} setting. A more general setting would include multivariate inputs and outputs, exogenous covariates, and probabilistic predictions. Although we anticipate similar conclusions, we do not consider it here.

%% file: plots/distrib-shift.tex
\begin{figure}[h!]
    \centering
    \begin{subfigure}[t]{0.3\textwidth}
        \centering
        \includegraphics[scale=0.3]{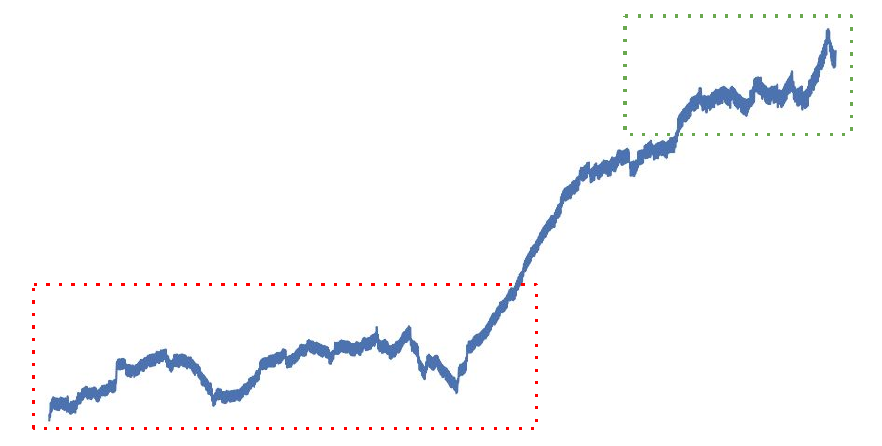}
        \caption{Temporal shift.}
        \label{fig:a}
    \end{subfigure}
    \hfill
    \begin{subfigure}[t]{0.3\textwidth}
        \centering
        \includegraphics[scale=0.2]{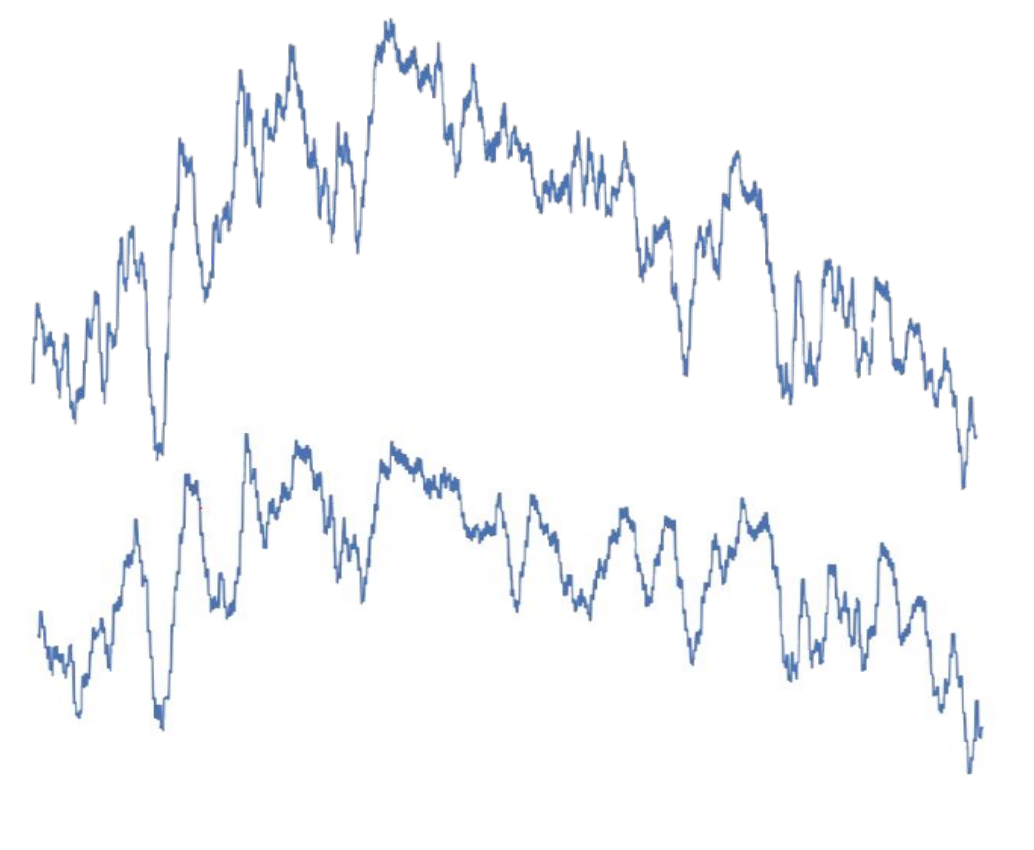}
        \caption{Spatial shift.}
        \label{fig:b}
    \end{subfigure}
    \hfill
    \begin{subfigure}[t]{0.3\textwidth}
        \centering
        \includegraphics[scale=0.3]{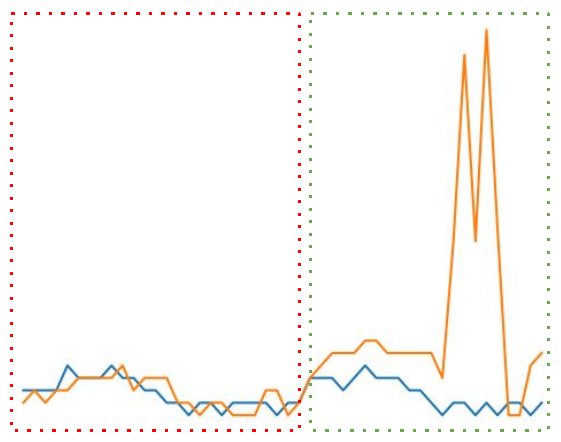}
        \caption{Conditional shift.}
        \label{fig:c}
    \end{subfigure}

    \caption{Illustration of the three distribution shifts: (a) temporal shift between training and test periods (rolling average of a \textsc{Traffic} sensor), (b) spatial shift between users (two \textsc{Solar} sensors), (c) conditional shift (different horizons for similar look-back windows, from one \textsc{Electricity} user).}
    \label{fig:shifts}
\end{figure}

%% file: 02--normalization-methods.tex

\section{Related Work} \label{related-content}

In this section, we describe the common normalization techniques for neural networks and review their limitations. More detailed mathematical formulations are provided in \cref{apx:norms}.

\paragraph{Normalization techniques in deep learning.} Early studies on neural networks optimization showed that having similarly distributed input features centered around the origin was beneficial for convergence and performance
\citep{DBLP:series/lncs/LeCunBOM12}. A common normalization technique is \textit{standard normalization}, which centers and reduces data distributions using the training data's statistics:
\begin{equation}
    \tilde{{x}}=\frac{{x}-{\mu}}{{\sigma}}, \quad \tilde{{y}}=\frac{{y}-{\mu}}{{\sigma}}
\end{equation}
where the statistics are computed empirically as:
\begin{equation}
    \mu=\frac{1}{N}\sum_{x\in \mathcal{D}}\frac{1}{L}\sum_{i=1}^L x_i \; \in \mathbb{R} \quad \textrm{and} \quad{{\sigma}}^2=\frac{1}{NL-1}\sum_{x\in \mathcal{D}}\sum_{i=1}^L( x_i-\mu)^2 \; >0.
\end{equation}


Later on, various normalization schemes were introduced between layers of deep neural networks to accelerate convergence, e.g. \textit{batch normalization} \citep{batchnorm} and \textit{layer normalization} \citep{layernorm}. Using the inputs' own statistics was introduced with \textit{instance normalization} \citep{instancenorm}. A thorough analysis of these methods can be found in \citep{beyondbatchnorm}.


\paragraph{Normalization for time series forecasting.} 
Normalization has long been used not only to facilitate optimization but also to mitigate time-series non-stationarity, as recognized early on, e.g. by \citet{adaptivenorm}.
Indeed, standard normalization cannot handle distribution shifts, since $(\mu,\sigma)$ no longer correspond to the correct statistics at inference time. Nevertheless, until recently, standard normalization---sometimes called \textit{standardization} or \textit{z-normalization}---has remained the default strategy \citep{informer,dain}. Inspired by the normalization techniques from other domains, \citet{revin} proposed \textit{Reversible Instance Normalization} for time series forecasting.

With RevIN, the prediction pipeline associated with a model $f_\theta: x \mapsto \hat{y}$ becomes:
\begin{equation}
\tilde{{x}}=\alpha\frac{{x}-{\mu}_x}{{\sigma_x}}+\beta, \quad 
\hat{y} = \sigma_x \left( \frac{f_\theta(\tilde{x}) - \beta}{\alpha} \right) + \mu_x \quad (\alpha,\beta \in \mathbb{R}),
\end{equation}
where the \textit{instance} $x$'s statistics are computed as:
\begin{equation}
\mu_x = \frac{1}{L}\sum_{i=1}^L x_i \;\in \mathbb{R}, \quad \sigma_x^2 = \frac{1}{L-1}\sum_{i=1}^L(x_i-\mu_x)^2\;>0.
\end{equation}

The pipeline can be decomposed as applying the following steps:
\begin{enumerate*}[(a)]
    \item apply instance normalization, i.e. standardize using the input statistics $\mu_x$ and $\sigma_x$;
    \item apply a learnable affine transformation with global parameters $\alpha$ and $\beta$;
    \item pass the transformed 
    input through the forecaster $f_{\theta}$;
    \item apply the inverse affine transformation; and
    \item denormalize using the inverse of the normalization.
\end{enumerate*} A diagram of the whole process is presented in \cref{fig:revin-process}.

\input{plots/revin-process}

By normalizing input windows using their own statistics, RevIN addresses temporal distribution shifts and, consequently, substantially  improves performance \citep{revin}. One key ingredient is the denormalization step that enables the model to make predictions at the correct offset and scale.

\paragraph{Extensions to RevIN.}  A few works have identified limitations of RevIN and proposed extensions or alternatives. In \textit{DAIN} \citep{dain}, per instance normalization is applied using learned scale and offset factors (more general functions of $x$ replacing the combination of $\mu_x,\sigma_x$ and $\alpha,\beta$), but no denormalization module is proposed. In \textit{Non-Stationary Transformers} \citep{nonstattransfo}, input statistics are reinjected into the model’s attention layers to mitigate the \textit{over-stationarization} effect, i.e. losing relevant information when normalizing. The authors also point out that $\alpha$ and $\beta$ in RevIN can be removed with no harm to the model. In \textit{Dish-TS} \citep{dishts}, a shift similar to our conditional shift is mentioned, and the authors propose learning both the normalization and denormalization factors as functions of the input, which requires careful regularization. In \citet{san}, normalization is applied to slices of the instance window, and future statistics are predicted based on past ones. In \citet{inflow}, a modern formulation of RevIN is developed using \textit{normalization flows} and \textit{bi-level optimization}, but essentially reverts to standard statistics with an additional affine layer learned on the validation data. 
Overall, the majority of these works have identified different limitations of RevIN which we summarize as follows: normalizing and denormalizing by $\mu_x,\sigma_x$ introduces a strong inductive bias, which mitigates certain distribution shifts but also limits models' capacity to learn a robust conditional distribution. Yet the proposed solutions are either incomplete \citep{dain}, model-specific \citep{nonstattransfo} or require additional complex training procedures \citep{dishts,san,inflow}. To our knowledge, no efficient and lightweight alternative has yet emerged. In this paper, we seek to properly identify the limitations of RevIN and draw perspectives on how to solve them.


%% file: plots/revin-process.tex
\begin{figure}[h!]
    \centering
    \vspace{-5mm}
    \includegraphics[width=\linewidth]{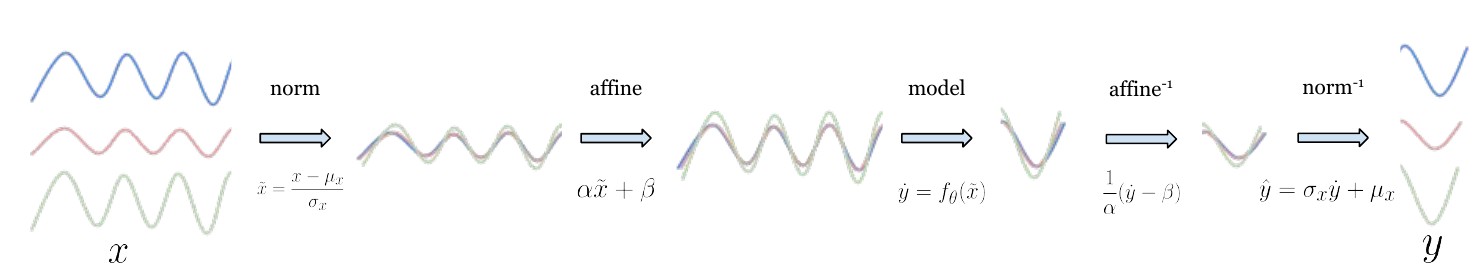}
    \vspace{-5mm}
    \caption{Illustration of the RevIN process on three synthetic examples.}
    \label{fig:revin-process}
\end{figure}

%% file: 03--experiments.tex
\section{Investigating RevIN} \label{ablation-studies}

\subsection{Quantitative Experiments} \label{quantitative}

In this section, we quantitatively evaluate the impact of the RevIN framework and its individual components, assessing whether each component contributes to improved generalization. 


\paragraph{Normalization strategies.} 
We compare several normalization strategies applied during training:
\begin{itemize}
    \item \textbf{Standard normalization:} Inputs are normalized using shared mean $\mu$ and standard deviation $\sigma$, computed empirically over all the training data.
    \item \textbf{RevIN:} Inputs are normalized independently based on the look-back window's statistics $\mu_x$ and $\sigma_x$, with an additional learnable affine transformation ($\alpha,\beta$). The output of the model is denormalized using the inverse of the normalization layer.
    \item \textbf{RevIN (w/o $\alpha,\beta$):} Same as RevIN but without the learnable affine transformation.
\end{itemize}

\paragraph{Training strategies.} For RevIN, we compare two training strategies:
\begin{itemize}
    \item \textbf{Standard backpropagation:} The training losses are computed between the denormalized predictions $\hat{y}$ and the ground truths $y$, which corresponds to the \textit{data space}.
    \item \textbf{Normalized backpropagation:} The training losses are computed on non-denormalized predictions and normalized ground-truths $\tilde{y}=\frac{y- \mu_x}{\sigma_x}$ (using the look-back's statistics), i.e. in the \textit{normalized space}. For RevIN with $\alpha,\beta$, this comes after the inverse $\textrm{affine}$ layer (see \cref{fig:revin-process}).
\end{itemize}

\paragraph{Datasets.} 
Experiments are conducted on three widely used real-world datasets—\textsc{Electricity}, \textsc{Solar}, and \textsc{Traffic}—as well as a controlled synthetic dataset (see \cref{apx:synthetic}). For each dataset, we consider multiple look-back ($L$) and horizon ($H$) configurations, denoted as $L$-$H$: short-term (168–24), medium-term (504–24 and 504–168), and long-term (504–504). Evaluation windows are sampled from unseen users and dates. Overall train-test split procedure is detailed in \cref{apx:protocol}.

\paragraph{Protocol.}  
To isolate the effect of normalization, all experiments are conducted under identical architectures and optimization settings. We use the \textsc{PatchTST} architecture \citep{patchtst} as the backbone neural forecaster. It is a transformer-based model that has achieved state-of-the-art results in long-horizon forecasting. For each normalization variant, we train models for 1200 epochs with a learning rate of $10^{-5}$, a batch size of $256$ and the Adam optimizer \citep{adam}. Each experiment is repeated across 5 random seeds, and we report the mean test MSE. Full details on experimental settings are provided in \cref{apx:protocol}. Results are presented in \cref{tab:main} (full tables in \cref{apx:results}).

\input{tables/main-table}

\paragraph{Results.}  \cref{tab:main} yields several observations. First, instance normalization is not uniformly best across tasks. Nevertheless, 
on average, it outperforms other normalization techniques on new dates and new users (see \cref{tab:all_in2} for results on new dates with identical users). This suggests that RevIN mitigates not only temporal shifts but also spatial ones, thus addressing both  challenges (i) and (ii) from \cref{intro}. Second, the additional affine layer (parameters $\alpha$ and $\beta$ ) is not beneficial in practice, and in particular, does not mitigate conditional shifts (challenge (iii)). We discuss the role of instance normalization and possible extensions to mitigate challenge (iii) in the next section. Third, although RevIN computes the loss on \emph{denormalized} predictions (i.e., after inverting the instance normalization), training via backpropagation in the normalized space yields better models. This holds even when evaluating with the non-normalized MSE, which we view as the more meaningful metric. While this observation may appear counter-intuitive, we hypothesize that models learned in the normalized space better generalize by attributing the same weight both to low-scale and high-scale instances. We note that recent works, e.g., \citet{moirai}, often train directly in the normalized space and omit the affine layer---referring to this variant simply as \textit{instance normalization}---but do not justify this design  choice. Our study shows these choices are indeed beneficial. \cref{fig:examples} shows example predictions for the various normalization strategies.


\subsection{Discussion} \label{qualitative}

\paragraph{Mitigating heterogeneity.} By applying instance normalization, we essentially project the data in a space with stationary first and second order statistics. Indeed, by linearity, we immediately have: $\mu(\tilde{x})=\beta$ and $\sigma(\tilde{x})=\alpha$ ($\mu(\tilde{x})=0$ and $\sigma(\tilde{x})=1$ when the affine layer is removed).
Thus, if we note $\mathcal{P}_{\textrm{training}}$ and  $\mathcal{P}_{\textrm{test}}$ the data distributions for the training and test periods, the heterogeneity RevIN addresses and actually mitigates is:
\begin{equation}
    \mathbb{E}_{x\sim \mathcal{P}_{\textrm{training}}}[\mu_x]\neq \mathbb{E}_{x\sim\mathcal{P}_{\textrm{test}}}[\mu_x] \quad \textrm{and} \quad \mathbb{E}_{x\sim \mathcal{P}_{\textrm{training}}}[\sigma_x]\neq \mathbb{E}_{x\sim\mathcal{P}_{\textrm{test}}}[\sigma_x].
\end{equation}

This simple definition of heterogeneity cannot fully address the distribution shift problem, since training and test distributions can be arbitrarily different. Indeed, \cref{apx:tsne} (t-SNE embeddings) and \cref{apx:distances} (distance metrics) show that instance normalization generally reduces heterogeneity, but neither completely nor consistently. In particular, on \textsc{Traffic}, which exhibits low temporal and spatial heterogeneity, instance normalization increases the distributions' distance. This would explain the degraded performance in \cref{tab:main}. Thus, addressing a more general definition of heterogeneity is advised to improve RevIN. For example, \citet{fan} normalize in the whole frequency domain, which is a promising direction.

\paragraph{Conditional shift.} \citet{revin} argue that the discrepancy between statistics of the look-back and horizon windows should remain approximately constant over the whole distribution of windows, and that their method allows the model to focus on learning the overall stationary behavior. This would be the case if the output affine layer were to learn such a fixed shift. In \cref{apx:min}, we show, on synthetic data, that carefully designed affine layers with RevIN can indeed learn meaningful shifts. However, we also show that real-world datasets rarely exhibit such fixed shifts, which may explain why the affine layer has little effect in our experiments. 
Moreover, by removing scale and offset from the input, RevIN discards potentially predictive context. A simple example of why this can be harmful is provided in \cref{apx:residual}. Overall, robustness to conditional changes likely requires retaining dependence on the original statistics, although doing so in an architecture-agnostic manner is not straightforward and ongoing research.

%% file: tables/main-table.tex
\begin{table}[h!]
\caption{Ablation of RevIN components on \textsc{PatchTST} (MSE on new dates and new users). Average improvements are expressed relative to the first column. BP stands for ``backpropagation".}
\vspace{0.2cm}
\centering
\scalebox{0.6}{
\begin{tabular}{lccccccc}
\toprule
& & & & \multicolumn{2}{c}{RevIN (w/o $\alpha,\beta$)} & \multicolumn{2}{c}{RevIN} \\
     \cmidrule(r){5-6} \cmidrule(r){7-8}
 & L-H &  None & Standard Normalization& \thead{Standard BP} & \thead{Normalized BP} & \thead{Standard BP} & \thead{Normalized BP} \\
\midrule
\multirow{4}{*}{Electricity} & 168-24 & 61.16 & 23.55 & 0.93 & \textbf{0.59} & 0.93 & \textbf{0.59} \\
 & 504-24 & 60.82 & 22.32 & 0.92 & \textbf{0.52} & 0.92 &\textbf{ 0.52} \\
 & 504-168 & 60.27 & 21.33 & 1.02 & \textbf{0.67} & 1.01 & \textbf{0.67} \\
 & 504-504 & 60.82 & 23.04 & 1.23 & \textbf{0.87} & 1.22 & \textbf{0.87} \\
\midrule
\multirow{4}{*}{Solar} & 168-24 & 2.74 & 2.35 & 1.31 & \textbf{1.30} & 1.31 & \textbf{1.30} \\
 & 504-24 & 2.74 & 2.20 & 1.28 & \textbf{1.23} & 1.28 &\textbf{ 1.23} \\
 & 504-168 & 2.54 & 3.24 & 1.97 & \textbf{1.94} & 1.97 & \textbf{1.94} \\
 & 504-504 & 2.52 & 3.59 & 2.13 & \textbf{2.06} & 2.13 & 2.07 \\
\midrule
\multirow{4}{*}{Traffic} & 168-24 & 14.44 & \textbf{6.51} & 7.30 & 7.27 & 7.30 & 7.27 \\
 & 504-24 & 19.60 & \textbf{6.02} & 6.77 & 6.72 & 6.77 & 6.72 \\
 & 504-168 & 18.75 & \textbf{6.87} & 8.13 & 8.09 & 8.13 & 8.09 \\
 & 504-504 & 17.87 & \textbf{6.99} & 8.29 & 8.26 & 8.28 & 8.25 \\
\midrule
\multirow{3}{*}{Synthetic} & 40-10 & $3.6\times10^{6}$ & $1.5\times10^{3}$ & \textbf{4.30} & \textbf{4.30} & \textbf{4.30} & \textbf{4.30} \\
 & 100-20 & $3.3\times10^{6}$ & $5.6\times10^{3}$ & \textbf{3.68} &\textbf{ 3.68} & \textbf{3.68} & \textbf{3.68} \\
 & 100-100 & $3.3\times10^{6}$ & $1.1\times10^{4}$ & \textbf{4.30} & \textbf{4.30} & \textbf{4.30} & \textbf{4.30} \\
\midrule
Improvements &  & 0.0 $\%$  & 6.97 $\%$   & 70.62 $\%$   & \textbf{71.29} $\%$   & 70.63 $\%$   & \textbf{71.29} $\%$   \\
\bottomrule
\end{tabular}}
\label{tab:main}
\end{table}

%% file: 04--conclusion.tex
\section{Conclusion} \label{conclu}

In this paper, we have investigated the role of instance normalization in time series forecasting. We have reviewed state of the art normalization strategies, in particular the popular \textit{Reversible Instance Normalization}. We have shown experimentally that input normalization and backpropagation in the normalized space are important components of the overall pipeline. However, we have also pointed out that the additional linear layer is not required, and that normalizing by instance on certain stationary datasets might be detrimental. More importantly, RevIN does not address all forms of heterogeneity. In particular, we emphasized the importance of modeling the input/output conditional distribution.

\clearpage

%% file: appendix.tex
\begin{appendices}
\crefalias{section}{appendix}

\section{Distribution shifts in Time Series Forecasting}
\label{apx:shifts}

In this section, we formalize the various distribution shifts from \cref{intro}. To do so, we must precisely define the data distributions.

In \textit{autoregressive} time series forecasting, input data points are temporal windows of the form $(x,y)\in \mathbb{R}^{d\times L}\times \mathbb{R}^{d\times H}$, where $d$ is the number of variates, $L$ is the look-back and $H$ the horizon size. In \textit{univariate} forecasting, we sample windows from a pool of users (or sensors) and dates.

We assume data is generated by an overall distribution $\mathcal{P}_{X,Y}$. For a subset of users $\mathcal{I}$ and period $\mathcal{T}$, we note the data distribution as $\mathcal{P}^{\mathcal{I}, \mathcal{T}}_{X,Y}$. Deep learning models aim to approximate overall $\mathcal{P}_{Y|X}$ by performing empirical risk minimization on a training dataset corresponding to users $\mathcal{I}_{train}$ and dates $\mathcal{T}_{train}$. It is then evaluated on a potentially distinct set of users $\mathcal{I}_{test}$ and future dates $\mathcal{T}_{test}$. When data is independent and identically distributed, the training and test distributions converge towards the ground-truth distribution $\mathcal{P}_{X,Y}$ as the sample size increases.

However, real-world data may exhibit various distributional shifts:

\begin{enumerate}[(i)]
    \item \textbf{Temporal shift:} $\exists \,\mathcal{I} \;\textrm{such that} \;\mathcal{P}^{\mathcal{I},\mathcal{T}_{train}}_{X} \neq \mathcal{P}^{\mathcal{I},\mathcal{T}_{test}}_{X}$

    \item \textbf{Spatial shift:} $\exists\, \mathcal{T}\;\textrm{such that} \;\mathcal{P}^{\mathcal{I}_{train},\mathcal{T}}_{X} \neq \mathcal{P}^{\mathcal{I}_{test},\mathcal{T}}_{X}$

    \item \textbf{Conditional shift:} $\exists\, \mathcal{I},  \mathcal{I}', \mathcal{T}, \mathcal{T}'\;\textrm{such that} \;\mathcal{P}^{\mathcal{I},\mathcal{T}}_{Y|X} \neq \mathcal{P}^{\mathcal{I}',\mathcal{T}'}_{Y|X}$
\end{enumerate}

Shifts (i) and (ii) characterize the input domain, whereas shift (iii) characterizes the expected output given the input, from a user to another or a time point to another. 
Less formally, for shifts (i) and (ii), the challenge is at inference, where the model will see data that may be very different from what it saw during training. For shift (iii), the challenge is that the expected prediction for identical inputs depends on the temporal and spatial context, which is problematic both for inference and training.

Other papers have identified similar distribution shifts in time series forecasting. For instance, \citet{dishts} call shift (i) the \textit{intra-space shift} and define an \textit{inter-space shift} as $\mathcal{P}_X\neq \mathcal{P}_{Y}$ (more specifically in terms of windows' statistics, e.g $\mu$ and $\sigma$). Our shift (iii) is more general, as it considers the conditional distribution of $\mathcal{P}_{Y|X}$, and is not considered uniform in time and space. In the domain adaptation and federated learning communities, shifts (i) and (ii) are sometimes refered to as \textit{covariate shifts}, and shift (iii) as \textit{concept drift} \citep{kairouz2021advances}.

Recently in the time series forecasting community, more exotic forms of heterogeneity have also been considered. For example, \citet{fan} tackles heterogeneity on the whole frequency domain, thus including amplitude and seasonality changes. In \citet{innerinstance}, heterogeneity within an instance itself is considered, via point-wise scaling.

\clearpage
\section{Normalization methods}
\label{apx:norms}

In this section, we formally define normalization methods from the deep learning community, applied to \textit{auto-regressive} time series forecasting. We consider general time series datasets with dimensions $(\text{users},\text{variates},\text{dates})$. We sample input windows $x$ of shape $(d,L)$ ($d$ variates and look-back window of size $L$). We note $x\sim \mathcal{P}_{train}$ when $x$ is sampled from the training data. When sampled in batch, we will use the notation $\underline{x}\in \mathbb{R}^{B\times d\times L}$.

Normalization consists in applying an affine transformation to $x$, such that the normalized input $\tilde{x}$ is distributed more favorably for the model. Essentially, it changes the functional space in which the models lives, in the hopes of better convergence or generalization. The majority of normalization strategies can be grouped in three families:
\begin{itemize}
    \item \textbf{Min-Max}: constrain values inside [0,1].
    $$\tilde{x}=\frac{x-m}{M-m} \quad \textrm{where} \quad m=\min[x], \; M=\max[x]$$
    \item \textbf{Relative}: divide by the mean.
    $$\tilde{x}=\frac{x}{\mu} \quad \textrm{where} \quad \mu=\mathbb{E}[x]$$
    \item \textbf{Standardization strategies}: center and reduce towards the unit ball.
    $$\tilde{x}=\frac{x-\mu}{\sigma} \quad \textrm{where} \quad \mu=\mathbb{E}[x], \; \sigma^2=\mathbb{V}[x]$$
    In time series, this method is also called \textit{z-normalization}.\\
    In practice, $\sigma$ is replaced by $\sigma+\epsilon$ for numerical stability.
\end{itemize}
The various normalization strategies vary in how the parameters $m,M,\mu,\sigma$ are precisely computed:
\begin{itemize}
    \item Global: on the whole training data ($\mathcal{I}_{train}$ and $\mathcal{T}_{train}$).
    \item Per-user: on the training dates ($\mathcal{I}_{train}$).
    \item Per instance: for each window $x$.
\end{itemize}

In all these cases, the parameters' dimensions  can also vary:
\begin{itemize}
    \item Scalar ($\mathbb{R})$: computed across variates and temporal dimensions.
    \item Per-variate ($\mathbb{R}^d$): computed across the temporal dimension only.
\end{itemize}

Global per-variate strategies are the most common in the literature, though we can find some per-user as well \citep{autoformer}. Variations of these strategies also exist, for example $\sigma$ can be computed as a mean (across users) of per-user standard deviations (across time). \textit{Adaptive normalization} \citep{adaptivenorm} uses a global Min-Max strategy applied to the detrended series (values divided by their moving average).

In the following, we define more precisely the more common standardization strategies.

\paragraph{Standard z-normalization} Global standardization with statistics computed on the training data:
$$\mu=\mathbb{E}_{x\sim\mathcal{P}_{\text{train}}}[x] \in \mathbb{R^d}, \; \sigma^2=\mathbb{V}_{x\sim\mathcal{P}_{\text{train}}}[x]\in \mathbb{R^d}$$
This is the default normalization strategy, usually applied to datasets upstream of any model training. Additionally, the following normalization strategies are commonly applied between layers of neural networks.

\paragraph{Batch normalization} \citep{batchnorm} Compute the batch's statistics:
$$\underline{\tilde{x}}=\gamma \frac{x-\mu_{\underline{x}}}{\sigma_{\underline{x}}}+\beta \quad \textrm{with} \quad \mu_{\underline{x}},\sigma_{\underline{x}}\in \mathbb{R}^d \quad \textrm{(computed over the batch)}$$
The linear coefficients $\gamma,\beta \in \mathbb{R}^d$ introduce a learnable flexibility into the normalization. At inference, estimates of $\mathbb{E}[\mu_{\underline{x}}],\mathbb{E}[\sigma_{\underline{x}}]$ are used, which were computed during training using moving averages. This strategy is biased if data is not iid.
\paragraph{Layer normalization} \citep{layernorm} Compute the instance's scalar statistics:
$$\tilde{x}=\gamma \frac{x-\mu_x}{\sigma_x}+\beta \quad \textrm{with} \quad \mu_{x}, \sigma_x \in \mathbb{R}$$

\paragraph{Instance Normalization} \citep{instancenorm} Compute the instance's per-variate statistics:
$$\tilde{x}=\gamma \frac{x-\mu_x}{\sigma_x}+\beta \quad \textrm{with} \quad \mu_{x}, \sigma_x \in \mathbb{R}^d$$

These last two normalization techniques introduced the idea of computing per-instance statistics, and only differ in dimensionality. In the computer vision community, \textit{group normalization} \citep{groupnorm} was also introduced which is a compromise between both strategies: statistics are computed per groups of channels (e.g $\mu_x \in \mathbb{R}^{d//g}$).

The following normalization techniques were specifically designed to normalize inputs for time series forecasting, but take inspiration from the aforementioned strategies, in particular from \textit{instance normalization}.

\paragraph{DAIN.} \citep{dain} Learn the scaling factors using linear layers:
$$\tilde{x}=\gamma (x-\alpha)\beta \quad \textrm{where} \quad \alpha=f_\alpha(\mu_x),\beta=f_\beta(\sigma_x), \gamma=f_\gamma(\mu_{(x-\alpha)\beta})$$
\paragraph{RevIN.} \citep{revin} Normalize by instance and denormalize symmetrically after the model:
$$\tilde{x}=\alpha\frac{x-\mu_x}{\sigma_x}+\beta \quad \textrm{and} \quad \hat{y}= \sigma_x\frac{f_\theta(\tilde{x})-\beta}{\alpha}+\mu_x$$
\paragraph{DishTS.} \citep{dishts} Learn both input and output scaling factors:
$$\tilde{x}=\frac{x-\varphi}{\xi} \quad \textrm{where} \quad \varphi,\xi=f_{in}(x)$$
$$\hat{y}= \xi'f_\theta(\tilde{x})+\varphi' \quad \textrm{where} \quad \varphi',\xi'=f_{out}(x) $$
Careful regularization on the horizon values is required during training to learn $f_{out}$.
\paragraph{SAN.} \citep{san} Normalize successive temporal slices $x^i$ of $x$, and learn to predict future slices' statistics as a function of previous slices' statistics:
$$\tilde{x}^i=\frac{x^i-\mu_{x^i}}{\sigma_{x^i}} \quad \textrm{and} \quad \hat{y}^i= \sigma_{x_i}'f_\theta(\tilde{x})^i+\mu_{x^i}'$$
$$\textrm{where} \quad \mu_{x^i}'=f_\mu(\mu_{x^i}-\mu_x), \;  \sigma_{x^i}'=f_\sigma(\sigma_{x^i}-\sigma_x)$$

As one can see on \cref{tab:citations}, only RevIN has really gained
attention in the time series forecasting research community, which suggests we have yet to find its successor.

\begin{table}[!h]
    \centering
    \caption{Citations of TSF normalization papers, according to Google Scholar (06/03/26)}  
    \vspace{4mm}
    \begin{tabular}{|c|c|}
        \toprule
        Normalization & Citations \\
        \midrule
        DAIN & 301  \\
        \midrule
        RevIN & 1322 \\
        \midrule        
        DishTS & 173 \\
        \midrule
        SAN & 142 \\
        \bottomrule
    \end{tabular}
    \label{tab:citations}
\end{table}

\clearpage
\section{Experimental settings}
\label{apx:protocol}

In this section, we describe the experimental settings for \cref{tab:main} and tables in \cref{apx:results}.

\paragraph{Data splits.} We apply a novel \textit{6-way split}, which accounts for both spatial and temporal dimensions. Indeed, time series datasets consists in successive numerical values measured by multiple sensors (or ``users", we call this the \textit{spatial} dimension). To properly train and evaluate a time serie model, we must split dates sequentially into three periods $\mathcal{T}_{train},\mathcal{T}_{valid},\mathcal{T}_{test}$, as well as two distinct subsets of users $\mathcal{I}_{in}, \mathcal{I}_{out}$. The second subset contains users not seen during training, but the model. This enables to build six different data distributions:
\begin{table}[!h]
    \centering
    \begin{tabular}{llcc}
    \toprule
    Split & Description & Users & Period \\
    \midrule
    Train  & Training distribution & $\mathcal{I}_{in}$ & $\mathcal{T}_{train}$ \\
    Valid1  & Validation (new period) & $\mathcal{I}_{in}$ & $\mathcal{T}_{valid}$ \\
    Valid2  & Validation (new users) & $\mathcal{I}_{out}$ & $\mathcal{T}_{train}$ \\
    Valid3  & Validation (new period \& new users) & $\mathcal{I}_{out}$ & $\mathcal{T}_{valid}$ \\
    Test1  & Testing (new period) & $\mathcal{I}_{in}$ & $\mathcal{T}_{test}$ \\
    Test2  & Testing (new period \& new users) & $\mathcal{I}_{out}$ & $\mathcal{T}_{test}$ \\
    \bottomrule
    \end{tabular}
    \caption{6-way dataset split}
    \label{tab:6waysplit}
\end{table}

\paragraph{Data processing.} Some time series have large amounts of missing values, which are set to $0$, leading to periods of entirely constant windows. Such windows are problematic because \begin{enumerate*}[(1)]
    \item they have 0 variance and consequently cause exploding normalized values.
    \item the switch from constant to non-constant is sudden not predictable by any model.
\end{enumerate*}
When such windows are identified, we have decided to remove them entirely. This stabilizes training and leads to better results overall. For \textsc{Electricity}, we have removed users which contained too many missing values (e.g users 57, 106, 127, 182 \& 298).

\paragraph{Data sampling.} Due to their two-dimensional nature, sampling windows from time series datasets can be done in multiple ways. We found that indexing by users and sampling random dates lead to smoother convergence.

\paragraph{Hyperparameters.} Due to the large number of models and settings, we avoided optimizing hyperparameters per-setting, and instead chose a set of parameters that allowed smooth convergence in most cases: 1200 epochs, learning rate of $10^{-5}$, batch size of $256$ (Adam optimizer). Each experiment is repeated across 5 random seeds. Exceptions (time and memory constraints): batch size of $64$ for settings $(504-168)$ and $(504-504)$, and only one seed for \textsc{Traffic} at those settings.

\paragraph{Losses.} Unless specified otherwise, we train and evaluate models using the mean-squared error (MSE). In certain settings, we use the ``nMSE" (normalized MSE), which corresponds to the MSE in the normalized space:
$$nMSE(\hat{y},y) = MSE(\tilde{\hat{y}},\tilde{y})= ||\frac{\hat{y}-\mu_x}{\sigma_x}-\frac{y-\mu_x}{\sigma_x}||_2^2 \quad \textrm{where } \hat{y} \textrm{ is the denormalized prediction}$$

\clearpage





\clearpage
\section{On the affine layer learning the conditional shift}
\label{apx:min}

By reading into RevIN's appendix, we understand the authors make the assumption that there exists a \textit{fixed} shift between output and input statistics:
\begin{equation}
    \label{eq:assumption}
    \exists \delta,\lambda,\forall(x,y):\quad \mu_y = \mu_x + \delta  \quad \textrm{and} \quad \sigma_y=\lambda \sigma_x
\end{equation}

They justify this by the fact that look-back window and horizon are close in time. While this could suggest that $(\delta_x,\lambda_x)\approx (0, 1)$ (and perhaps only when $H\approx L$), it has no reason to imply that $\delta_x,\lambda_x$ are constant with respect to $x$.

Also, the goal of RevIN is to allow the model to map stationary distributions, in particular to have $\mathbb{E}[\tilde{x}],\mathbb{V}[\tilde{x}],\mathbb{E}[\tilde{y}], \mathbb{V}[\tilde{y}]$ not dependent on $x$. Using instance normalization, we already have $\mathbb{E}[\tilde{x}]=\beta,\mathbb{V}[\tilde{x}]=\alpha^2$. Yet for $\tilde{y}$, we have:

\begin{equation}
    \mathbb{E}[\tilde{y}]=\frac{\mu_y-\mu_x}{\sigma_x}\quad \textrm{and} \quad \mathbb{\sigma}[\tilde{y}]=\frac{\sigma_y}{\sigma_x}
\end{equation}

Thus, we'd better make the following assumptions:

\begin{equation}
    \label{eq:assumption_bis}
    \exists \delta,\lambda,\forall(x,y):\quad 
    \frac{\mu_y-\mu_x}{\sigma_x}=\delta \quad \textrm{and} \quad \frac{\sigma_y}{\sigma_x}=\gamma
\end{equation}

We call this assumption \textit{modulation stationarity} (and the associated constants the \textit{modulations}). From now on, we will note $z\sim (\mu,\sigma)$ to indicate that window $z$ has mean $\mu$ and standard deviation $\sigma$. In case of modulation stationarity, we have $\tilde{y} \sim (\delta,\lambda)$.

The authors from RevIN claim that using their method, a model can easily learn the modulations and focus on the stationary mapping. This lead us to wonder if the affine layer in the denormalization block could learn the statistical shift, thus allowing the model to be truly stationary (i.e. map from a given distribution to the same distribution). Essentially, can $(\alpha,\beta)$ learn $(\delta,\gamma)$ alone?

Unfortunately, this cannot be the case. Indeed, an input $x$ undergoes the following transformations:
$$x\sim (\mu_x,\sigma_x)\rightarrow \tilde{x}\sim (\beta,\alpha) \rightarrow f_\theta(\tilde{x}) \rightarrow \frac{f_\theta(\tilde{x})-\beta}{\alpha}:=\dot{y}$$

If we have modulation stationarity and the model is correctly fitted, then we'd expect the non-denormalized prediction to be distributed like $\tilde{y}$, i.e. $\dot{y}\sim(\delta,\gamma)$. Thus $f_\theta(x)\sim (\beta+\alpha\delta, \alpha\lambda)$. So the internal model would have to learn the $(\delta,\gamma)$ shift anyway.



We've shown that we cannot have a model with RevIN be unbiased as well as fully stationary. We argue that for the model to focus on the stationary behavior, the denormalization layer must not be symmetric to the normalization layer. We propose a simple modification to the RevIN framework, using an additional affine layer at denormalization:
$$ \tilde{x} = \gamma \frac{x-\mu_x}{\sigma_x}+\nu \quad \longrightarrow \quad \hat{y} =  \sigma_x  (\alpha\frac{f_\theta(\tilde{x}) - \nu}{\gamma} + \beta) + \mu_x$$

We further allow $(\alpha,\beta, \eta, \nu)$ to be conditioned to a given spatial and temporal context. In particular, for a given set of users $\mathcal{I}$, we apply the following initialization:
\begin{equation}
    \label{eq:alpha}
    (\gamma_{\mathcal{I}},\nu_{\mathcal{I}})=(1,0), \quad 
    (\beta_{\mathcal{I}}, \alpha_{\mathcal{I}} )= \mathbb{E}_{\mathcal{T}_{train}, \mathcal{I}}[\delta_x,\gamma_x]
\end{equation}

\begin{figure}
    \centering
    \includegraphics[width=0.7\linewidth]{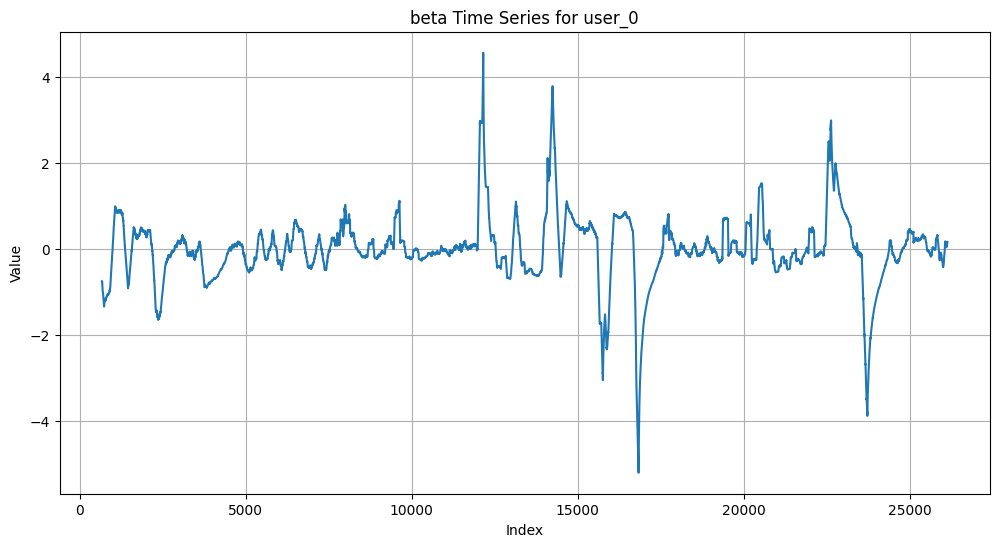}
    \caption{Values of $\delta$ for a given user in time. Modulation stationarity does not hold.}
    \label{fig:beta_time}
\end{figure}

We call this method \textit{cmIN}, for \textit{per-cluster modulated instance normalization}). We built a synthetic dataset containing two clusters of users with modulation stationarity (see \cref{apx:synthetic}). On \cref{tab:cmin}, we can see that personalizing non-symmetric layers leads to an improvement over RevIN, even more so if we initialize the output parameters as per \cref{eq:alpha}. Unfortunately, cmIN in its current form doesn't work on real-world datasets, mainly because modulation stationarity \textit{does not} hold in practice (see \cref{fig:beta_time}). Online tuning or meta-learning of cmIN are possible working directions.

\begin{table}[h!]
\caption{Results of cmIN (on PatchTST) for the synthetic dataset (nMSE, Test1 split).}
\vspace{0.2cm}
\centering
    \scalebox{0.8}{
    \begin{tabular}{lccccc}
    \toprule
    & L-H & Instance & RevIN & cmIN & cmIN(init)\\
    
    \midrule
    \multirow{3}{*}{Synthetic} & 40-10 & 427.42 & 427.30 & \underline{425.17} & \textbf{356.03} \\
     & 100-20 & 366.02 & 365.36 & \underline{364.70} & \textbf{328.54} \\
     & 100-100 & 425.38 & 425.22 & \underline{424.97} & \textbf{346.45} \\
    \midrule
    Improvements &  & 0.0 \%  & 0.08 \%  & \underline{0.33 \%}  & \textbf{15.16 \%}  \\
    \bottomrule
    \end{tabular}}
\label{tab:cmin}
\end{table}


\begin{figure}
    \centering
    \includegraphics[width=0.5\linewidth]{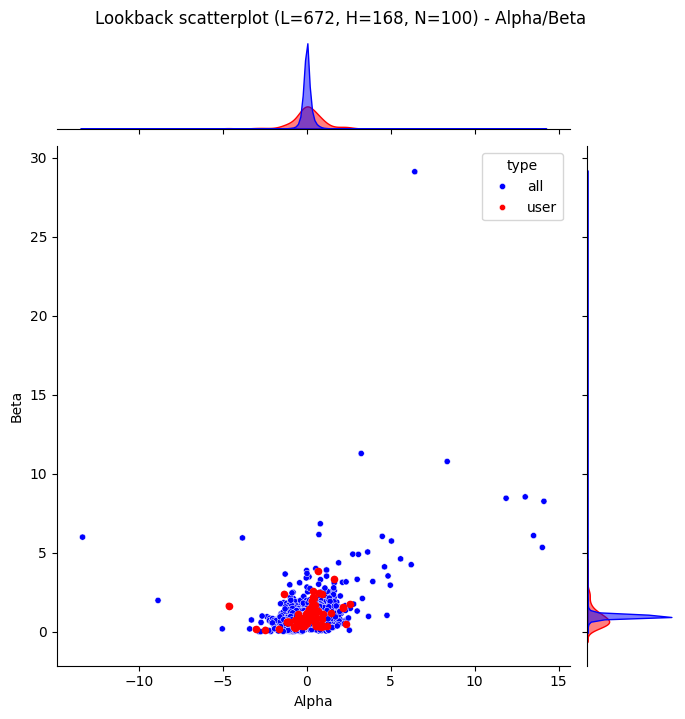}
    \caption{Distribution of sampled $(\delta,\lambda)$. In red from a single user, in blue from the whole set of users. Modulation stationarity does not hold.}
    \label{fig:gammas_plot}
\end{figure}

\clearpage
\section{On reintegrating statistics in the model's prediction}
\label{apx:residual}

In this section, we discuss the theoretical justifications for reintegrating statistics into the model's internals. Instance normalization projects inputs into a quotient space which is scale and offset invariant:
$$x\sim x' \Leftrightarrow \tilde{x}=\tilde{x}' $$
This means two series that only vary in scale and offset become identical. Yet such two equivalent windows do not necessarily have identical (denormalized) horizons:
$$x\sim x' \nRightarrow y = y'$$

When modeling $p(y|x)=p(x^{\textrm{horizon}}|x^{\textrm{look-back}})$, the goal of normalization is to factorize the distribution as follows \citep{inflow}:
$$p (y|x)=p(y|\tilde{y})\,p(\tilde{y}|\tilde{x})\,p(\tilde{x}|x)$$
Where $\tilde{y},\tilde{x}$ live in a stationary space that the model can easily map. In particular, it means that $p(\tilde{y},\tilde{x})$ is not context-dependant, i.e. it is the same for any user and period. RevIN's assumption is that $\exists \,\alpha,\beta \,\textrm{ such that } \,\tilde{y}=\frac{y-\mu_x}{\sigma_x}, \tilde{x}=\alpha\frac{x-\mu_x}{\sigma}+\beta$ satisfies this.

\begin{figure}[!h]
    \centering
    \includegraphics[width=0.5\linewidth]{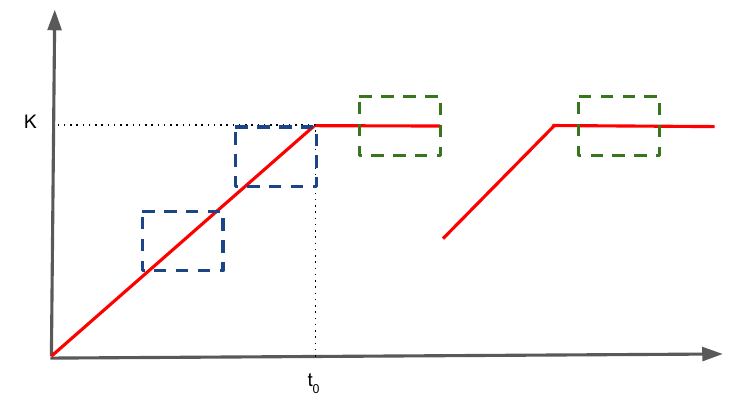}
    \caption{Example of a signal with saturation. A model with instance normalization cannot distinguish both blue windows nor both green windows. Yet the expected outputs are different for each.}
    \label{fig:saturation}
\end{figure}

Yet, we recall that the exact factorization is:
$$p(y|x)=p(y|\tilde{y},x)\,p(\tilde{y}|\tilde{x},x)\,p(\tilde{x}|x)$$
In particular, $\tilde{y}$ generally depends on $x$. In particular, it may depend on $\mu_x$. On \cref{fig:saturation}, we show a simple example where $\tilde{y}$ under instance normalization does depend on $\mu_x$, because there is a saturation effect.

With RevIN, we essentially assume:
\begin{itemize}
    \item $p(\tilde{y}|\tilde{x},x)=p(\tilde{y}|\tilde{x})$
    \item $p(y|\tilde{y},x)=p(y|\tilde{y},\mu_x,\sigma_x)$
\end{itemize}

In \citet{nonstattransfo}, the first assumption is softened by reintegrating the statistics in the models' internals, specifically at the attention layers. Yet finding how to reintegrate input statistics within any model architecture is yet to be found.

For the second assumption, softer inductive biases would be to find novel denormalization layers of the form: $\hat{y}=g\big(f_\theta(\tilde{x}),\mu_x,\sigma_x)$, and even weaker: $\hat{y}=g\big(f_\theta(\tilde{x}),x\big)$. These remain to be found, though we can mention attempts by \citet{dishts} and \citet{san}).

\clearpage
\section{Synthetic dataset}
\label{apx:synthetic}

Our synthetic dataset is built to experiment on a \textit{modulation stationary} setting (cf \cref{apx:min}). We build two clusters of users with sinusoidal time series. For each cluster, we add a constant horizon / look-back shift. An individual is characterized by a cyclicity $T \in \mathbb{N^*}$, an amplitude $A \in \mathbb{R}_+$, a scale and offset $a,b \in \mathbb{R}$, and some gaussian noise controlled by $\sigma>0$. The individual's time series of length $|\mathcal{T}|$ is then:
$$\forall t \in \mathcal{T}, \,X_t=at+b+A\sin(\frac{2\pi}{T}t)+ \epsilon_t \quad \quad  \big(\epsilon_t\sim \mathcal{N}(0,\sigma)\big)$$

\begin{figure}[!h]
    \centering
    \includegraphics[width=0.6\linewidth]{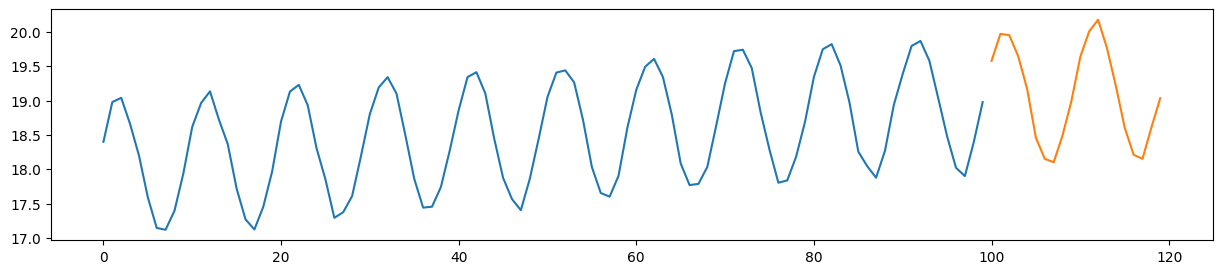}
    \includegraphics[width=0.6\linewidth]{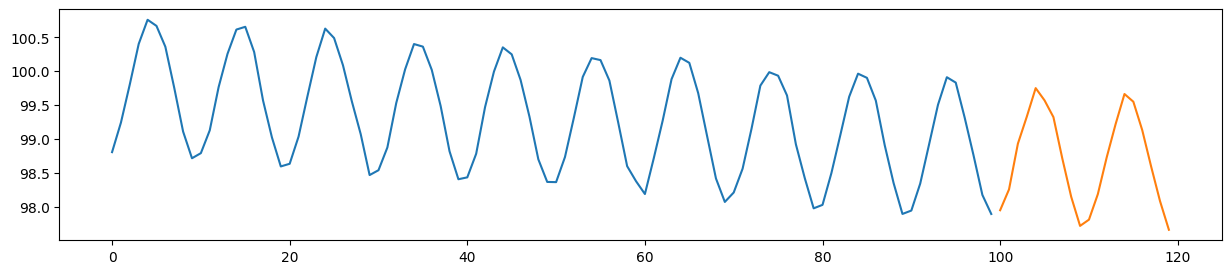}
    \caption{Example synthetic windows. Up: cluster 1 ($a=10^{-1}, b=10\pm1$), Down: cluster 2 ($a=-10^{-1}, b=100\pm10$). For all, we set $A=1, T=10, \sigma=5*10^{-2}$.}
    \label{fig:sim_eg}
\end{figure}

In this setting, we can compute the exact values for $(\delta,\lambda)$, which allows us to initialize modulations of cmIN (cf. \cref{apx:min}). For example, for $L=336$ and $H=48$ we get:
\begin{itemize}
    \item $\delta=\frac{\mu_y-\mu_x}{\sigma_x+\epsilon}=\frac{a\frac{L+H}{2}}{\sqrt{a^2L^2/12+1/2+\sigma^2}+\epsilon}=\frac{60a}{\sqrt{0.5025+\frac{100^2}{12}a^2}+10^{-6}}\approx_{a\pm0.01} \pm0.784$
    \item $\lambda=\frac{\sigma_y}{\sigma_x+\epsilon}=\frac{\sqrt{a^2H^2/12+1/2+\sigma^2}}{\sqrt{a^2L^2/12+1/2+\sigma^2}+\epsilon}=\frac{\sqrt{0.5025+\frac{100}{12}a^2}}{\sqrt{0.5025+\frac{100^2}{12}a^2}+10^{-6}}\approx_{a\pm0.01} 0.929$
\end{itemize}

And indeed, as per \cref{tab:cmin}, initializing cmIN at these values (recomputed each time empirically on the training split), we greatly improve performance.

\begin{figure}[!h]
    \centering
    \includegraphics[width=0.49\linewidth]{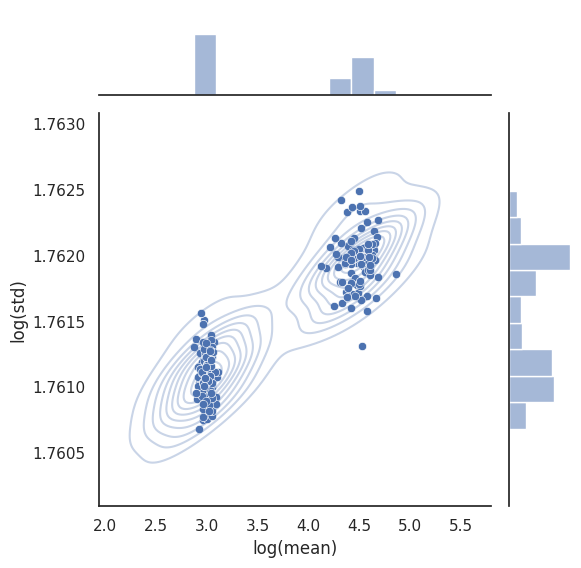}
    \includegraphics[width=0.49\linewidth]{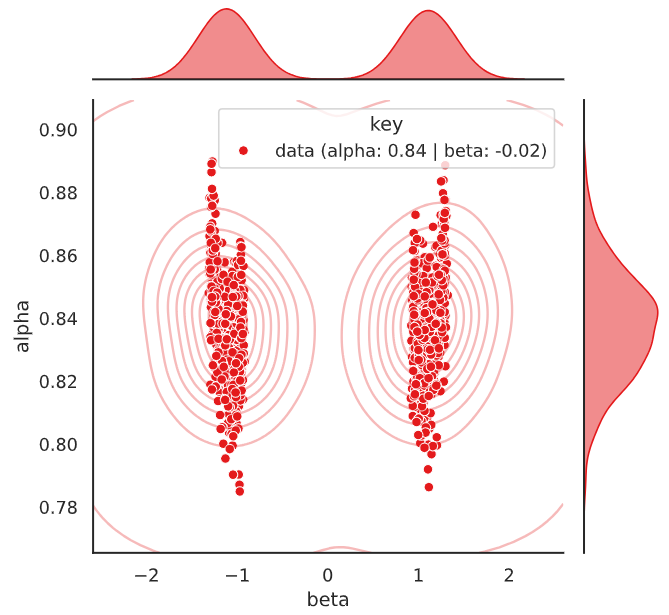}
    \caption{\textsc{Synthetic} dataset's statistics. We notice the two distinct clusters. Individuals are recreated randomly at each run.}
    \label{fig:synth_stats}
\end{figure}

\clearpage
\section{Full results}
\label{apx:results}
In this section, we include all quantitative results from the experiment of \cref{quantitative}. 

\input{tables/results}

\clearpage

\section{tSNE}
\label{apx:tsne}

In this section, we analyze the effect of instance normalization using \textit{t-SNE} \citep{tsne}. We sample windows from distinct distributions and project the features into a common latent space using \textit{t-SNE} embeddings. We consider four datasets: \textsc{Electricity}, \textsc{Solar}, \textsc{Traffic}, \textsc{Synthetic} (see \cref{apx:synthetic}). As well as three settings:
\begin{itemize}
    \item \textbf{Temporal shift}: inputs from training and test periods, for the same users (\textit{Train}, \textit{Test1})
    \item \textbf{Spatial shift}: inputs from training period, for two distinct sets of users (\textit{Train}, \textit{Valid2})
    \item \textbf{Overall shift}: inputs \& outputs from training and new users \& period (\textit{Train}, \textit{Test2})
\end{itemize}

Overall, we can see that instance normalization groups distinct data points together, but different input distributions (in time \& space) do not align entirely. In fact, instance normalization seems to increase distribution-wise distance in certain settings. More importantly, we see that there a regions of examples at inference which the model will have never seen during training. This suggests that RevIN cannot mitigate challenges (i) and (ii) entirely, let alone challenge (iii).

\input{plots/tsne}

\clearpage
\section{Distances}
\label{apx:distances}

In this section, we look into the distances between distributions to analyze the effect of instance normalization. For empirical distributions, a common choice for the distance metric is the \textit{Maximum Mean Discrepency}: 
\begin{equation*}
        d_{\text{MMD}}^2(P, Q)
        = \mathbb{E}_{x,x'\sim P}[k(x,x')]
        + \mathbb{E}_{y,y'\sim Q}[k(y,y')]
        - 2\,\mathbb{E}_{x\sim P,\,y\sim Q}[k(x,y)]
    \end{equation*}
where $k$ is a chosen kernel function. With kernel \( k(x,y) = -\|x - y\| \), we obtain the \textit{energy distance}. This distance can be computed in multiple distribution spaces. We chose four variants:
\begin{itemize}
    \item Whole inputs: $X = \textrm{(lookback)}\in \mathbb{R}^L$
    \item Whole windows: $X = \textrm{(lookback, horizon)}\in \mathbb{R}^{L+H}$ 
    \item Window statistics: $X=(\mu_x,\sigma_x) \in \mathbb{R}^2$
    \item  Modulations: $X=(\lambda_x,\delta_x) \in \mathbb{R}^2$
\end{itemize}

We evaluate two distribution shifts: Temporal (Train/Test1) and Spatial (Train/Valid2). In \cref{tab:distances_energy_a}, we see that the distances after normalization are not $0$ (thus the distributions not identical). On \cref{tab:distances_energy_c}, we can see that instance normalization does reduce the statistical heterogeneity entirely, as expected (see discussion in \cref{qualitative}). Nevertheless, it is not always the stronger method. For instance, standard normalization on \textsc{Electricity} reduces the distance between distributions more. Also, we see that for \textsc{Traffic}, both normalization methods increase the temporal gap.

\input{plots/distances}

\clearpage

\section{Additional figures}
\label{apx:additional}

\begin{figure}[!h]
    \centering
    \includegraphics[width=0.9\linewidth]{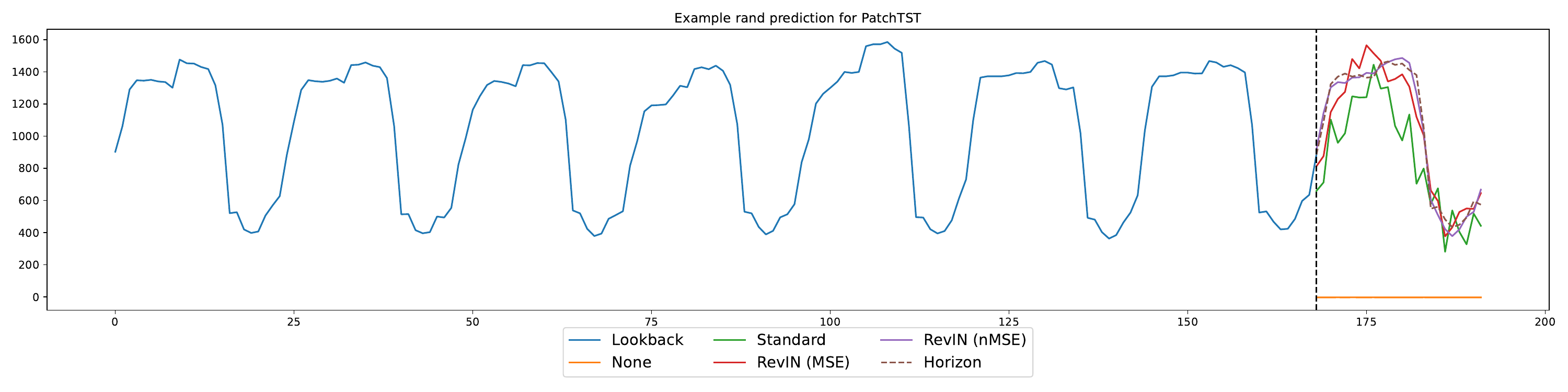}
    \caption{Example prediction from \textsc{Electricity} $(L-H)=(168-24)$. Without normalization, \textsc{PatchTST} does not converge. RevIN(nMSE) corresponds to RevIN trained with normalized backpropagation. It fits the ground-truth more smoothly than RevIN(MSE).}
    \label{fig:examples}
\end{figure}

\begin{figure}[!h]
    \centering
    \includegraphics[width=0.49\linewidth]{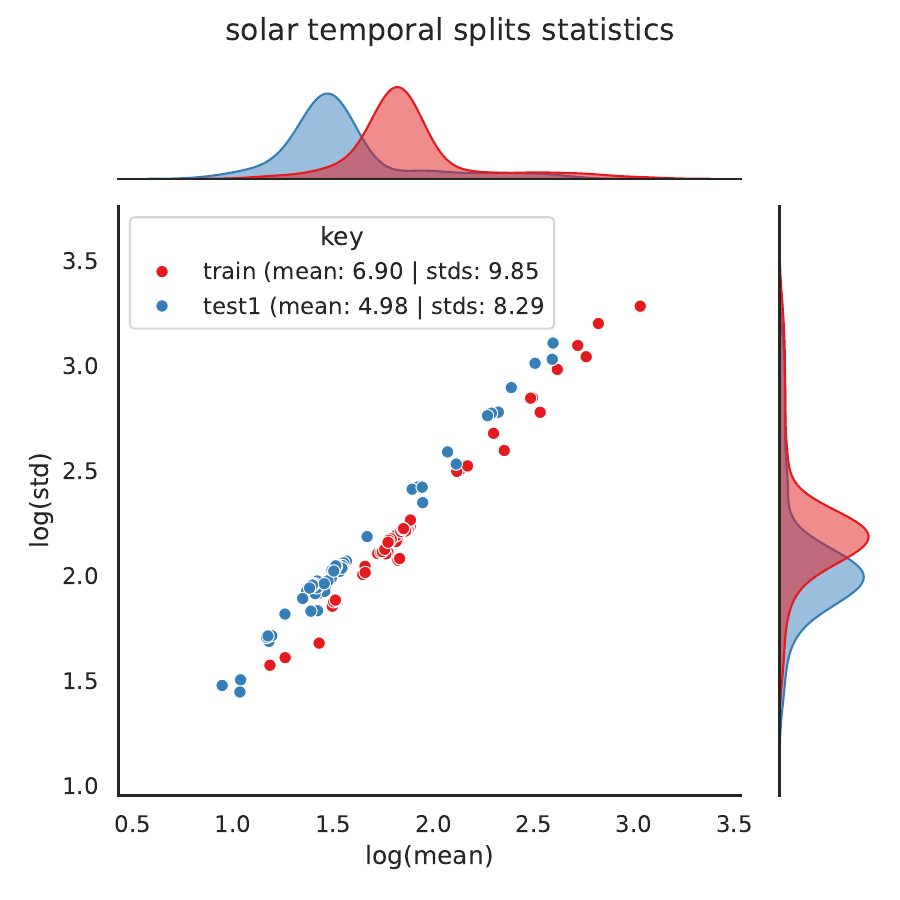}
    \includegraphics[width=0.49\linewidth]{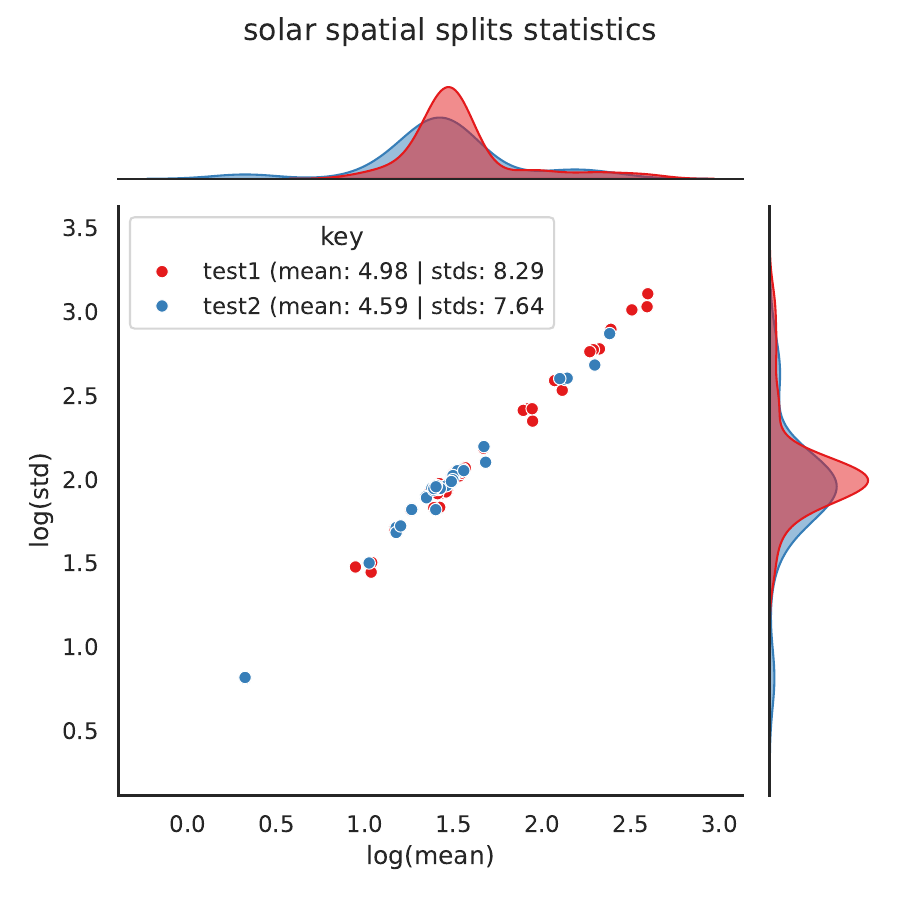}
    \caption{\textsc{Solar} sensors' statistics (Left: Train and Test1, Right: Test 1 and Test2).}
    \label{fig:solar_stats}
\end{figure}

\begin{figure}[!h]
    \centering
    \includegraphics[width=0.49\linewidth]{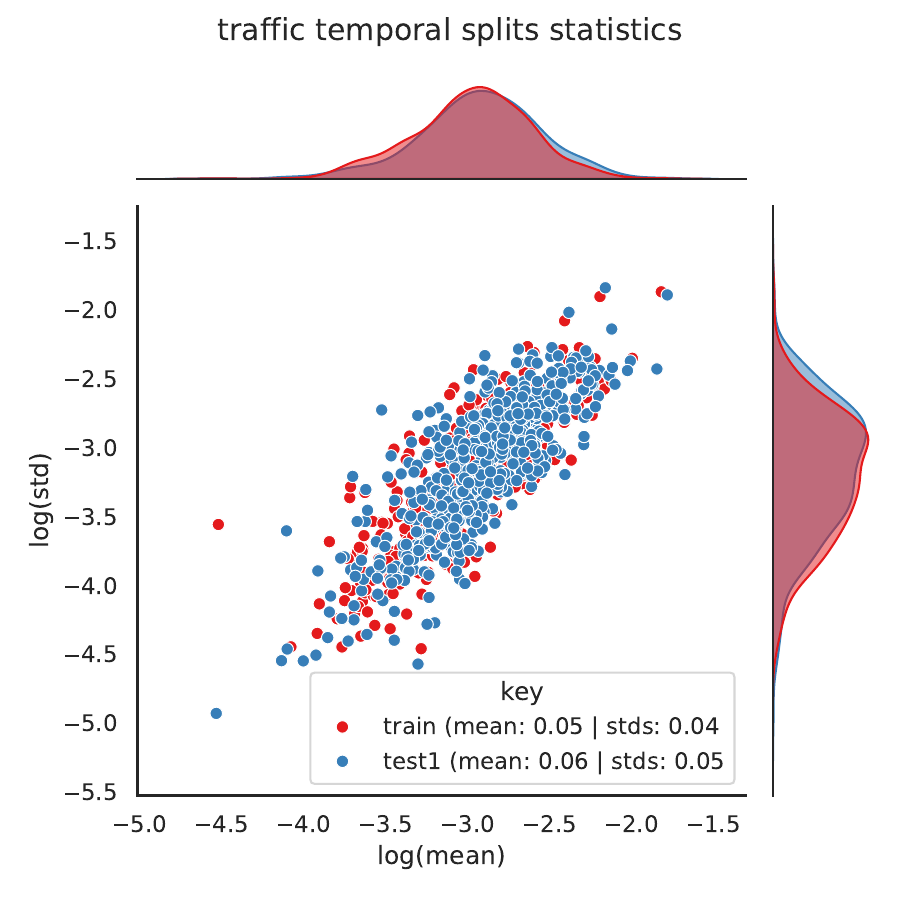}
    \includegraphics[width=0.49\linewidth]{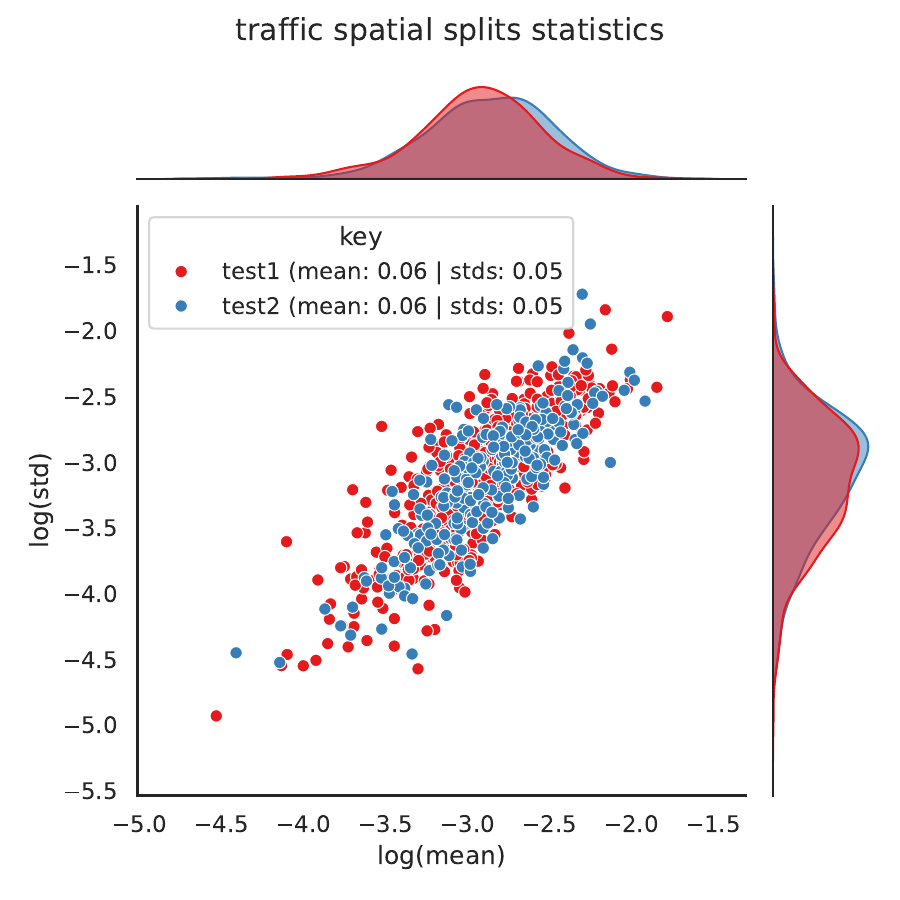}
    \caption{\textsc{Traffic} sensors' statistics (Left: Train and Test1, Right: Test 1 and Test2).}
    \label{fig:traffic_stats}
\end{figure}

\end{appendices}

%% file: tables/results.tex
\begin{table}[h!]
\caption{Ablation of RevIN components on \textsc{PatchTST} (MSE test1 results).}
\vspace{0.2cm}
\centering
\scalebox{0.6}{
\begin{tabular}{lccccccc}
\toprule
& &  &  & \multicolumn{2}{c}{RevIN (w/o $\alpha,\beta$)} & \multicolumn{2}{c}{RevIN} \\
     \cmidrule(r){5-6} \cmidrule(r){7-8}
 & L-H &  None & Standard Normalization& \thead{Standard BP} & \thead{Normalized BP} & \thead{Standard BP} & \thead{Normalized BP} \\
\midrule
\multirow{4}{*}{Electricity} & 168-24 & 349.83 & 20.56 & 8.52 & \textbf{4.91} & 8.52 & 4.92 \\
 & 504-24 & 348.14 & 19.68 & 8.36 & \textbf{4.09} & 8.34 & 4.09 \\
 & 504-168 & 347.21 & 28.28 & 11.00 &\textbf{ 7.09} & 10.94 & \textbf{7.09} \\
 & 504-504 & 358.70 & 33.18 & 16.70 & 12.54 & 16.65 & \textbf{12.53} \\
\midrule
\multirow{4}{*}{Solar} & 168-24 & 3.05 & 2.30 & 1.47 & \textbf{1.46} & 1.47 & 1.47 \\
 & 504-24 & 3.12 & 2.16 & 1.44 & \textbf{1.38} & 1.44 & 1.38 \\
 & 504-168 & 2.81 & 3.19 & 2.22 & \textbf{2.18} & 2.22 & 2.18 \\
 & 504-504 & 2.82 & 3.55 & 2.40 & \textbf{2.33} & 2.40 & \textbf{2.33 }\\
\midrule
\multirow{4}{*}{Traffic} & 168-24 & 15.74 & \textbf{6.96} & 8.10 & 8.07 & 8.10 & 8.07 \\
 & 504-24 & 19.62 & \textbf{6.44} & 7.52 & 7.47 & 7.52 & 7.47 \\
 & 504-168 & 19.85 & \textbf{7.10} & 8.39 & 8.36 & 8.38 & 8.35 \\
 & 504-504 & 19.10 & \textbf{7.17} & 8.49 & 8.47 & 8.48 & 8.47 \\
\midrule
\multirow{3}{*}{Synthetic} & 40-10 & 3.4$\times10^{6}$ & 1.5$\times10^3$  & 4.28 & 4.27 & 4.28 & \textbf{4.27} \\
 & 100-20 & 3.2$\times10^{6}$ & 5.3$\times10^{3}$ & 3.66 & 3.66 & 3.65 & \textbf{3.65} \\
 & 100-100 & 3.2$\times10^{6}$ & 9.8$\times10^{3}$ & 4.25 & 4.25 & \textbf{4.25} & 4.25 \\
\midrule
Improvements &  & 0.0 $\%$  & 62.32 $\%$  & 70.1 $\%$  & \textbf{70.84} $\%$  & 70.11 $\%$  & 70.84 $\%$  \\
\bottomrule
\end{tabular}}
\label{tab:all_in1}
\end{table}

\begin{table}[h!]
\caption{Ablation of RevIN components on \textsc{PatchTST} (MSE test2 results).}
\vspace{0.2cm}
\centering
\scalebox{0.6}{
\begin{tabular}{lccccccc}
\toprule
& &  &  & \multicolumn{2}{c}{RevIN (w/o $\alpha,\beta$)} & \multicolumn{2}{c}{RevIN} \\
     \cmidrule(r){5-6} \cmidrule(r){7-8}
 & L-H &  None & Standard Normalization& \thead{Standard BP} & \thead{Normalized BP} & \thead{Standard BP} & \thead{Normalized BP} \\
\midrule
\multirow{4}{*}{Electricity} & 168-24 & 61.16 & 23.55 & 0.93 & \textbf{0.59} & 0.93 & 0.59 \\
 & 504-24 & 60.82 & 22.32 & 0.92 & \textbf{0.52} & 0.92 & 0.52 \\
 & 504-168 & 60.27 & 21.33 & 1.02 & 0.67 & 1.01 & \textbf{0.67} \\
 & 504-504 & 60.82 & 23.04 & 1.23 & 0.87 & 1.22 & \textbf{0.87} \\
\midrule
\multirow{4}{*}{Solar} & 168-24 & 2.74 & 2.35 & 1.31 & \textbf{1.30} & 1.31 & 1.30 \\
 & 504-24 & 2.74 & 2.20 & 1.28 & \textbf{1.23} & 1.28 & 1.23 \\
 & 504-168 & 2.54 & 3.24 & 1.97 & \textbf{1.94} & 1.97 & 1.94 \\
 & 504-504 & 2.52 & 3.59 & 2.13 & \textbf{2.06} & 2.13 & 2.07 \\
\midrule
\multirow{4}{*}{Traffic} & 168-24 & 14.44 & \textbf{6.51} & 7.30 & 7.27 & 7.30 & 7.27 \\
 & 504-24 & 19.60 & \textbf{6.02} & 6.77 & 6.72 & 6.77 & 6.72 \\
 & 504-168 & 18.75 & \textbf{6.87} & 8.13 & 8.09 & 8.13 & 8.09 \\
 & 504-504 & 17.87 & \textbf{6.99} & 8.29 & 8.26 & 8.28 & 8.25 \\
\midrule
\multirow{3}{*}{Synthetic} & 40-10 & 3.6$\times10^6$ & 1.5$\times10^3$ & 4.30 & 4.30 & 4.30 & \textbf{4.30} \\
 & 100-20 & 3.3$\times10^6$ & 5.6$\times10^3$ & 3.68 & 3.68 & 3.68 & \textbf{3.68} \\
 & 100-100 & 3.3$\times10^6$ & 1.0$\times10^4$ & 4.30 & 4.30 & \textbf{4.30} & 4.30 \\
\midrule
Improvements &  & 0.0 $\%$  & 6.97 $\%$  & 70.62 $\%$  & \textbf{71.29} $\%$  & 70.63 $\%$  & 71.29 $\%$  \\
\bottomrule
\end{tabular}}
\label{tab:all_in2}
\end{table}

\begin{table}[h!]
\caption{Ablation of RevIN components on \textsc{PatchTST} (nMSE test1 results).}
\vspace{0.2cm}
\centering
\scalebox{0.6}{
\begin{tabular}{lccccccc}
\toprule
& &  &  & \multicolumn{2}{c}{RevIN (w/o $\alpha,\beta$)} & \multicolumn{2}{c}{RevIN} \\
     \cmidrule(r){5-6} \cmidrule(r){7-8}
 & L-H &  None & Standard Normalization& \thead{Standard BP} & \thead{Normalized BP} & \thead{Standard BP} & \thead{Normalized BP} \\
\midrule
\multirow{4}{*}{Electricity} & 168-24 & 49.28 & 4.7$\times10^{11}$ & 0.44 & \textbf{0.28} & 0.44 & 0.28 \\
 & 504-24 & 39.93 & 4.8$\times10^{11}$ & 0.34 & \textbf{0.19} & 0.34 & 0.19 \\
 & 504-168 & 52.56 & 5.4$\times10^{11}$ & 0.42 & 0.28 & 0.42 & \textbf{0.28} \\
 & 504-504 & 47.95 & 2.2$\times10^11$ & 0.60 & 0.45 & 0.60 & \textbf{0.45} \\
\midrule
\multirow{4}{*}{Solar} & 168-24 & 1.72 & 120.62 & 1.00 & \textbf{0.86} & 1.00 & 0.86 \\
 & 504-24 & 0.60 & 38.68 & 0.31 & \textbf{0.28} & 0.31 & 0.28 \\
 & 504-168 & 0.56 & 48.56 & 0.40 & 0.37 & 0.40 & \textbf{0.37} \\
 & 504-504 & 0.49 & 50.18 & 0.40 & \textbf{0.37} & 0.40 & 0.37 \\
\midrule
\multirow{4}{*}{Traffic} & 168-24 & 0.72 & \textbf{0.00} & 0.32 & 0.31 & 0.32 & 0.31 \\
 & 504-24 & 1.05 & \textbf{0.00} & 0.27 & 0.26 & 0.27 & 0.26 \\
 & 504-168 & 1.36 & \textbf{0.00} & 0.31 & 0.30 & 0.31 & 0.30 \\
 & 504-504 & 1.07 & \textbf{0.00} & 0.33 & 0.33 & 0.33 & 0.33 \\
\midrule
\multirow{3}{*}{Synthetic} & 40-10 & 6.5$\times10^3$ & 2.2$\times10^3$ & 0.01 & 0.01 & 0.01 & \textbf{0.01} \\
 & 100-20 & 5.4$\times10^3$ & 9.03 & 0.01 & 0.01 & 0.01 & \textbf{0.01} \\
 & 100-100 & 5.4$\times10^3$ & 16.48 & 0.01 & 0.01 & \textbf{0.01} & 0.01 \\
\midrule
Improvements &  & 0.0 $\%$  & 2.4$\times10^11$ $\%$  & 74.11 $\%$  & \textbf{75.8} $\%$  & 74.11 $\%$  & 75.8 $\%$  \\
\bottomrule
\end{tabular}}
\label{tab:all_in3}
\end{table}

\begin{table}[h!]
\caption{Ablation of RevIN components on \textsc{PatchTST} (nMSE test2 results).}
\vspace{0.2cm}
\centering
\scalebox{0.6}{
\begin{tabular}{lccccccc}
\toprule
& &  &  & \multicolumn{2}{c}{RevIN (w/o $\alpha,\beta$)} & \multicolumn{2}{c}{RevIN} \\
     \cmidrule(r){5-6} \cmidrule(r){7-8}
 & L-H &  None & Standard Normalization& \thead{Standard BP} & \thead{Normalized BP} & \thead{Standard BP} & \thead{Normalized BP} \\
\midrule
\multirow{4}{*}{Electricity} & 168-24 & 38.27 & 2.4$\times10^{12}$ & 0.48 & \textbf{0.30} & 0.48 & 0.30 \\
 & 504-24 & 32.35 & 1.0$\times10^{12}$ & 0.36 & \textbf{0.19} & 0.36 & 0.19 \\
 & 504-168 & 35.99 & 4.4$\times10^{11}$ & 0.43 & 0.27 & 0.43 & \textbf{0.27} \\
 & 504-504 & 37.31 & 6.1$\times10^{11}$ & 0.61 & 0.43 & 0.61 & \textbf{0.43} \\
\midrule
\multirow{4}{*}{Solar} & 168-24 & 1.88 & 132.03 & 1.09 & \textbf{0.93} & 1.09 & 0.93 \\
 & 504-24 & 0.61 & 38.45 & 0.30 & \textbf{0.28} & 0.30 & 0.28 \\
 & 504-168 & 0.58 & 48.56 & 0.39 & 0.37 & 0.39 & \textbf{0.37} \\
 & 504-504 & 0.53 & 50.17 & 0.40 & \textbf{0.37} & 0.40 & 0.37 \\
\midrule
\multirow{4}{*}{Traffic} & 168-24 & 0.70 & \textbf{0.00} & 0.31 & 0.30 & 0.31 & 0.30 \\
 & 504-24 & 1.03 & \textbf{0.00} & 0.26 & 0.26 & 0.26 & 0.26 \\
 & 504-168 & 1.15 & \textbf{0.00} & 0.30 & 0.30 & 0.30 & 0.30 \\
 & 504-504 & 0.94 & \textbf{0.00} & 0.33 & 0.33 & 0.33 & 0.33 \\
\midrule
\multirow{3}{*}{Synthetic} & 40-10 & 6.3$\times10^3$ & 2.2$\times10^3$ & 0.01 & 0.01 & 0.01 & \textbf{0.01} \\
 & 100-20 & 5.6$\times10^3$ & 9.39 & 0.01 & 0.01 & 0.01 & \textbf{0.01} \\
 & 100-100 & 5.6$\times10^3$ & 17.22 & 0.01 & 0.01 & \textbf{0.01} & 0.01 \\
\midrule
Improvements &  & 0.0 $\%$  & 7.2$\times10^{11}$ $\%$  & 74.1 $\%$  & \textbf{75.77} $\%$  & 74.1 $\%$  & 75.77 $\%$  \\
\bottomrule
\end{tabular}}
\label{tab:all_in4}
\end{table}

\begin{table}[h!]
\caption{Ablation of RevIN components on \textsc{DLinear} (MSE test1 results).}
\vspace{0.2cm}
\centering
\scalebox{0.6}{
\begin{tabular}{lccccccc}
\toprule
& &  &  & \multicolumn{2}{c}{RevIN (w/o $\alpha,\beta$)} & \multicolumn{2}{c}{RevIN} \\
     \cmidrule(r){5-6} \cmidrule(r){7-8}
 & L-H &  None & Standard Normalization& \thead{Standard BP} & \thead{Normalized BP} & \thead{Standard BP} & \thead{Normalized BP} \\
\midrule
\multirow{4}{*}{Electricity} & 168-24 & 88.15 & 78.28 & 45.01 & 22.04 & 45.01 & \textbf{22.04} \\
 & 504-24 & 20.92 & 18.55 & 17.00 & \textbf{11.18} & 17.00 & 11.19 \\
 & 504-168 & 17.82 & 15.72 & 16.04 & \textbf{8.89} & 16.04 & 8.90 \\
 & 504-504 & 23.06 & 20.34 & 21.09 & \textbf{14.54} & 21.08 & 14.63 \\
\midrule
\multirow{4}{*}{Solar} & 168-24 & 5.43 & 6.29 & 3.83 & \textbf{3.74} & 3.83 & 3.75 \\
 & 504-24 & 3.34 & 4.34 & 3.00 & \textbf{2.94} & 3.00 & 2.94 \\
 & 504-168 & 3.07 & 4.14 & 2.93 & 2.93 & \textbf{2.93} & 2.94 \\
 & 504-504 & 3.19 & 4.32 & 3.09 & 3.12 & \textbf{3.09} & 3.13 \\
\midrule
\multirow{4}{*}{Traffic} & 168-24 & 16.69 & \textbf{11.59} & 13.28 & 13.36 & 13.28 & 13.37 \\
 & 504-24 & 11.55 & \textbf{9.06} & 10.51 & 10.43 & 10.51 & 10.44 \\
 & 504-168 & 9.57 & \textbf{7.96} & 9.43 & 9.29 & 9.43 & 9.29 \\
 & 504-504 & 9.64 & \textbf{8.01} & 9.50 & 9.38 & 9.50 & 9.38 \\
\midrule
\multirow{3}{*}{Synthetic} & 40-10 & 3.2$\times10^{6}$ & 1.3$\times10^{6}$ & 402.27 & 402.30 & \textbf{401.73} & 401.77 \\
 & 100-20 & 1.2$\times10^{6}$ & 1.3$\times10^{6}$ & 368.62 & 368.57 & 368.52 & \textbf{368.47} \\
 & 100-100 & 9.6$\times10^{5}$ & 9.7$\times10^{5}$ & 921.86 & 922.00 & \textbf{921.81} & 921.95 \\
\midrule
Improvements &  & 0.0 $\%$  & 4.82 $\%$  & 30.98 $\%$  & \textbf{39.52} $\%$  & 31.01 $\%$  & 39.44 $\%$  \\
\bottomrule
\end{tabular}}
\label{tab:all_in5}
\end{table}

\begin{table}[h!]
\caption{Ablation of RevIN components on \textsc{DLinear} (MSE test2 results).}
\vspace{0.2cm}
\centering
\scalebox{0.6}{
\begin{tabular}{lccccccc}
\toprule
& &  &  & \multicolumn{2}{c}{RevIN (w/o $\alpha,\beta$)} & \multicolumn{2}{c}{RevIN} \\
     \cmidrule(r){5-6} \cmidrule(r){7-8}
 & L-H &  None & Standard Normalization& \thead{Standard BP} & \thead{Normalized BP} & \thead{Standard BP} & \thead{Normalized BP} \\
\midrule
\multirow{4}{*}{Electricity} & 168-24 & 9.14 & 60.96 & 2.74 & 1.49 & 2.74 & \textbf{1.49} \\
 & 504-24 & 1.70 & 12.44 & 1.43 & \textbf{0.96} & 1.43 & 0.96 \\
 & 504-168 & 1.39 & 10.67 & 1.29 & \textbf{0.83} & 1.29 & 0.83 \\
 & 504-504 & 1.57 & 11.72 & 1.43 & \textbf{1.02} & 1.43 & 1.03 \\
\midrule
\multirow{4}{*}{Solar} & 168-24 & 4.84 & 6.32 & 3.44 & \textbf{3.36} & 3.44 & 3.36 \\
 & 504-24 & 2.99 & 4.43 & 2.69 & \textbf{2.63} & 2.68 & 2.63 \\
 & 504-168 & 2.73 & 4.21 & 2.61 & 2.61 & \textbf{2.61} & 2.62 \\
 & 504-504 & 2.82 & 4.36 & 2.73 & 2.75 & \textbf{2.73} & 2.76 \\
\midrule
\multirow{4}{*}{Traffic} & 168-24 & 15.43 & \textbf{10.90} & 12.01 & 12.08 & 12.01 & 12.08 \\
 & 504-24 & 10.50 & \textbf{8.47} & 9.45 & 9.38 & 9.45 & 9.38 \\
 & 504-168 & 9.21 & \textbf{7.65} & 9.08 & 8.94 & 9.08 & 8.94 \\
 & 504-504 & 9.36 & \textbf{7.76} & 9.22 & 9.09 & 9.22 & 9.09 \\
\midrule
\multirow{3}{*}{Synthetic} & 40-10 & 3.3$\times10^{6}$ & 1.3$\times10^{6}$ & 402.60 & 402.64 & \textbf{402.07} & 402.11 \\
 & 100-20 & 1.3$\times10^{6}$ & 1.3$\times10^{6}$ & 368.69 & 368.64 & 368.59 & \textbf{368.54} \\
 & 100-100 & 1.0$\times10^{6}$ & 1.0$\times10^6$ & 921.71 & 921.92 & \textbf{921.67} & 921.87 \\
\midrule
Improvements &  & 0.0 $\%$  & -996.47 $\%$  & 31.86 $\%$  & \textbf{40.66} $\%$  & 31.88 $\%$  & 40.59 $\%$  \\
\bottomrule
\end{tabular}}
\label{tab:all_in6}
\end{table}

\begin{table}[h!]
\caption{Ablation of RevIN components on \textsc{DLinear} (nMSE test1 results).}
\vspace{0.2cm}
\centering
\scalebox{0.6}{
\begin{tabular}{lccccccc}
\toprule
& &  &  & \multicolumn{2}{c}{RevIN (w/o $\alpha,\beta$)} & \multicolumn{2}{c}{RevIN} \\
     \cmidrule(r){5-6} \cmidrule(r){7-8}
 & L-H &  None & Standard Normalization& \thead{Standard BP} & \thead{Normalized BP} & \thead{Standard BP} & \thead{Normalized BP} \\
\midrule
\multirow{4}{*}{Electricity} & 168-24 & 5.78 & 4.1$\times10^{12}$ & 0.83 & 0.57 & 0.83 & \textbf{0.57} \\
 & 504-24 & 0.60 & 2.9$\times10^{11}$ & 0.54 & 0.34 & 0.54 & \textbf{0.34} \\
 & 504-168 & 0.63 & 4.2$\times10^{10}$ & 0.56 & \textbf{0.34} & 0.56 & 0.34 \\
 & 504-504 & 0.79 & 4.8$\times10^{10}$ & 0.72 & \textbf{0.50} & 0.72 & 0.50 \\
\midrule
\multirow{4}{*}{Solar} & 168-24 & 1.98 & 237.45 & 1.67 & 1.64 & 1.67 & \textbf{1.64} \\
 & 504-24 & 0.69 & 71.44 & 0.62 & 0.61 & 0.62 & \textbf{0.60} \\
 & 504-168 & 0.58 & 62.52 & 0.55 & 0.54 & 0.55 & \textbf{0.54} \\
 & 504-504 & 0.56 & 61.02 & 0.54 & 0.53 & 0.54 & \textbf{0.53} \\
\midrule
\multirow{4}{*}{Traffic} & 168-24 & 0.90 & \textbf{0.00} & 0.48 & 0.48 & 0.48 & 0.48 \\
 & 504-24 & 0.43 & \textbf{0.00} & 0.36 & 0.36 & 0.36 & 0.36 \\
 & 504-168 & 0.35 & \textbf{0.00} & 0.34 & 0.34 & 0.34 & 0.34 \\
 & 504-504 & 0.37 & \textbf{0.00} & 0.36 & 0.36 & 0.36 & 0.36 \\
\midrule
\multirow{3}{*}{Synthetic} & 40-10 & 6.0$\times10^3$ & 1.9$\times10^{6}$ & 0.76 & 0.76 & 0.76 & \textbf{0.76} \\
 & 100-20 & 2.1$\times10^3$ & 2.1$\times10^3$ & 0.62 & 0.62 & 0.62 & \textbf{0.62} \\
 & 100-100 & 1.6$\times10^3$ & 1.6$\times10^3$ & 1.56 & 1.56 & 1.56 & \textbf{1.56} \\
\midrule
Improvements &  & 0.0 $\%$  & -8.6$\times10^{12}$ $\%$  & 34.29 $\%$  & 41.79 $\%$  & 34.28 $\%$  & \textbf{41.8} $\%$  \\
\bottomrule
\end{tabular}}
\label{tab:all_in7}
\end{table}

\begin{table}[h!]
\caption{Ablation of RevIN components on \textsc{DLinear} (nMSE test2 results).}
\vspace{0.2cm}
\centering
\scalebox{0.6}{
\begin{tabular}{lccccccc}
\toprule
& &  &  & \multicolumn{2}{c}{RevIN (w/o $\alpha,\beta$)} & \multicolumn{2}{c}{RevIN} \\
     \cmidrule(r){5-6} \cmidrule(r){7-8}
 & L-H &  None & Standard Normalization& \thead{Standard BP} & \thead{Normalized BP} & \thead{Standard BP} & \thead{Normalized BP} \\
\midrule
\multirow{4}{*}{Electricity} & 168-24 & 4.92 & 2.2$\times10^{12}$ & 0.87 & 0.60 & 0.87 & \textbf{0.60} \\
 & 504-24 & 0.63 & 7.4$\times10^{10}$ & 0.57 & 0.34 & 0.57 & \textbf{0.34} \\
 & 504-168 & 0.64 & 3.1$\times10^{10}$  & 0.58 & \textbf{0.32} & 0.58 & 0.32 \\
 & 504-504 & 0.82 & 4.5$\times10^{10}$ & 0.74 & \textbf{0.49} & 0.74 & 0.49 \\
\midrule
\multirow{4}{*}{Solar} & 168-24 & 2.11 & 254.75 & 1.80 & 1.77 & 1.80 & \textbf{1.77} \\
 & 504-24 & 0.69 & 72.22 & 0.62 & 0.61 & 0.62 & \textbf{0.60} \\
 & 504-168 & 0.58 & 62.65 & 0.55 & 0.54 & 0.55 & \textbf{0.54} \\
 & 504-504 & 0.56 & 60.92 & 0.54 & 0.53 & 0.54 & \textbf{0.53} \\
\midrule
\multirow{4}{*}{Traffic} & 168-24 & 0.88 & \textbf{0.00} & 0.47 & 0.47 & 0.47 & 0.47 \\
 & 504-24 & 0.42 & \textbf{0.00} & 0.35 & 0.35 & 0.35 & 0.35 \\
 & 504-168 & 0.34 & \textbf{0.00} & 0.33 & 0.33 & 0.33 & 0.33 \\
 & 504-504 & 0.37 & \textbf{0.00} & 0.36 & 0.36 & 0.36 & 0.36 \\
\midrule
\multirow{3}{*}{Synthetic} & 40-10 & 6.2$\times10^3$ & 2.0$\times10^{6}$ & 0.76 & 0.76 & 0.76 & \textbf{0.76} \\
 & 100-20 & 2.2$\times10^3$ & 2.2$\times10^3$ & 0.62 & 0.62 & 0.62 & \textbf{0.62} \\
 & 100-100 & 1.7$\times10^3$ & 1.7$\times10^3$ & 1.56 & 1.56 & 1.56 & \textbf{1.56} \\
\midrule
Improvements &  & 0.0 $\%$  & -4.3$\times10^{12}\%$  & 33.84 $\%$  & 42.11 $\%$  & 33.83 $\%$  & \textbf{42.12} $\%$  \\
\bottomrule
\end{tabular}}
\label{tab:all_in8}
\end{table}

%% file: plots/tsne.tex
\begin{figure}[!h]
    \centering
    \includegraphics[width=0.6\linewidth]{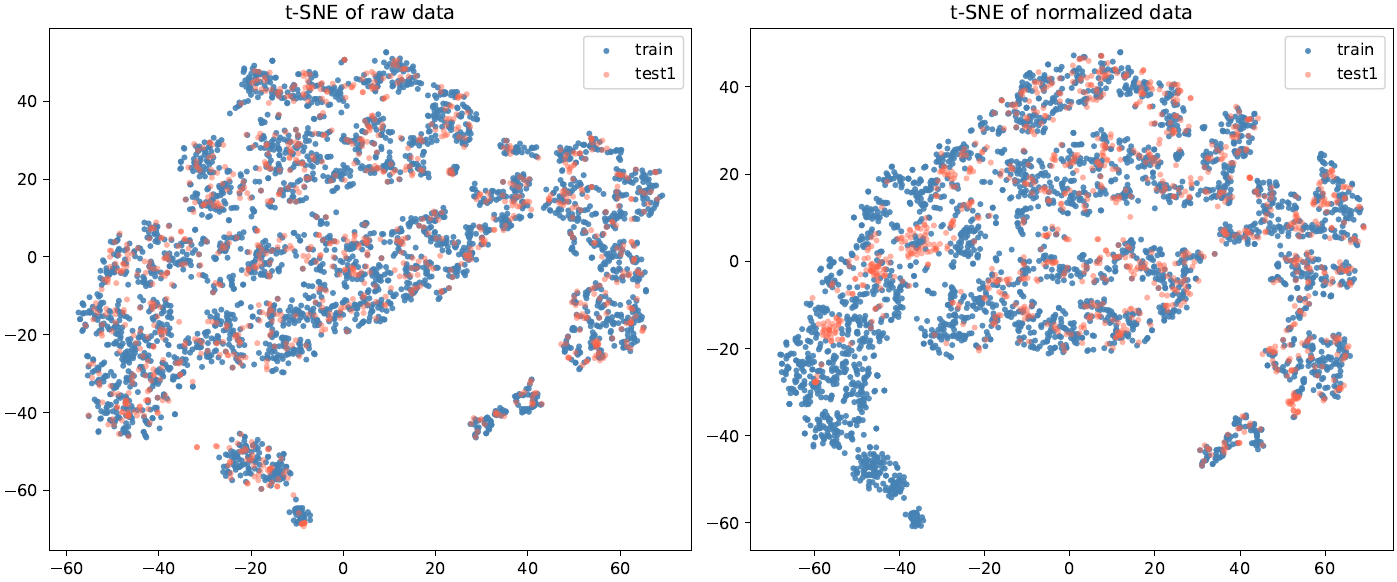}
    \caption{\textsc{Electricity}, temporal setting}
    \label{fig:tsne_ecl_temp}
\end{figure}

\begin{figure}[!h]
    \centering
    \includegraphics[width=0.6\linewidth]{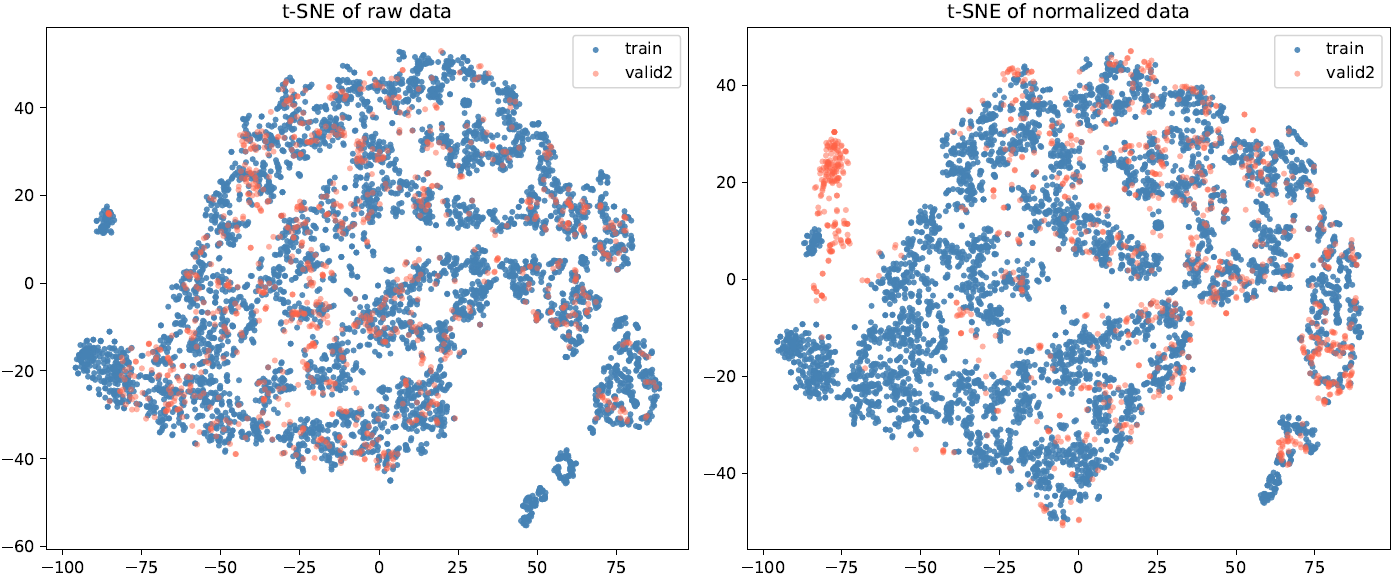}
    \caption{\textsc{Electricity}, spatial setting}
    \label{fig:tsne_ecl_spat}
\end{figure}

\begin{figure}[!h]
    \centering
    \includegraphics[width=0.6\linewidth]{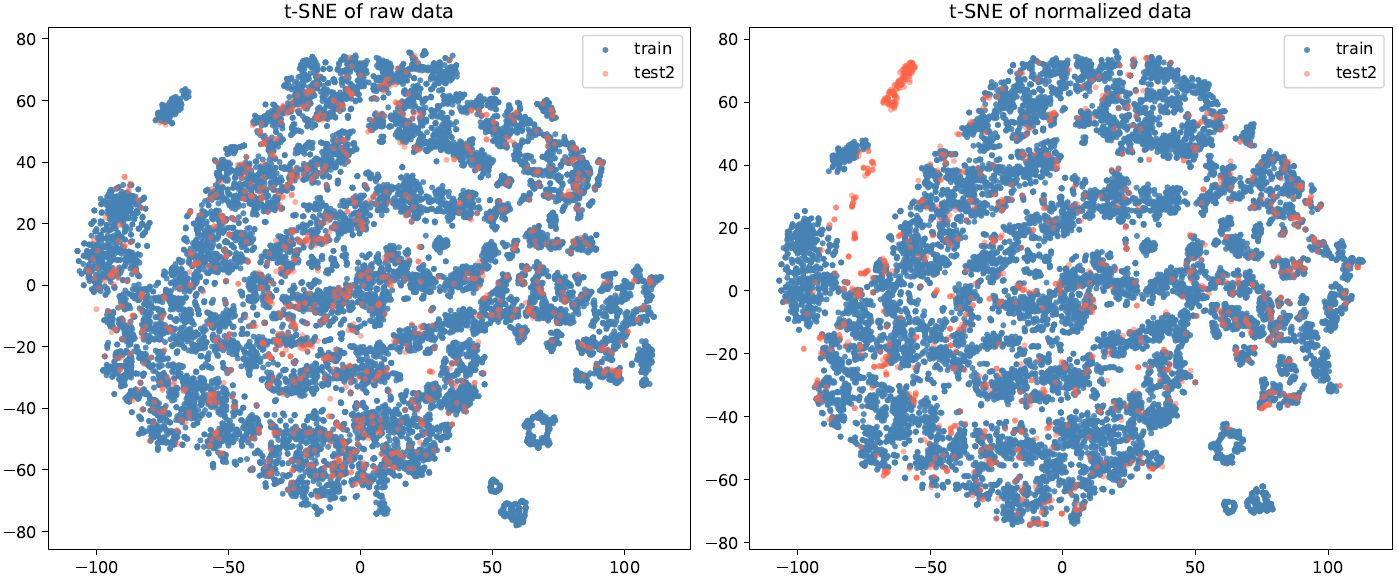}
    \caption{\textsc{Electricity}, Overall setting}
    \label{fig:tsne_ecl_cond}
\end{figure}

\begin{figure}[!h]
    \centering
    \includegraphics[width=0.6\linewidth]{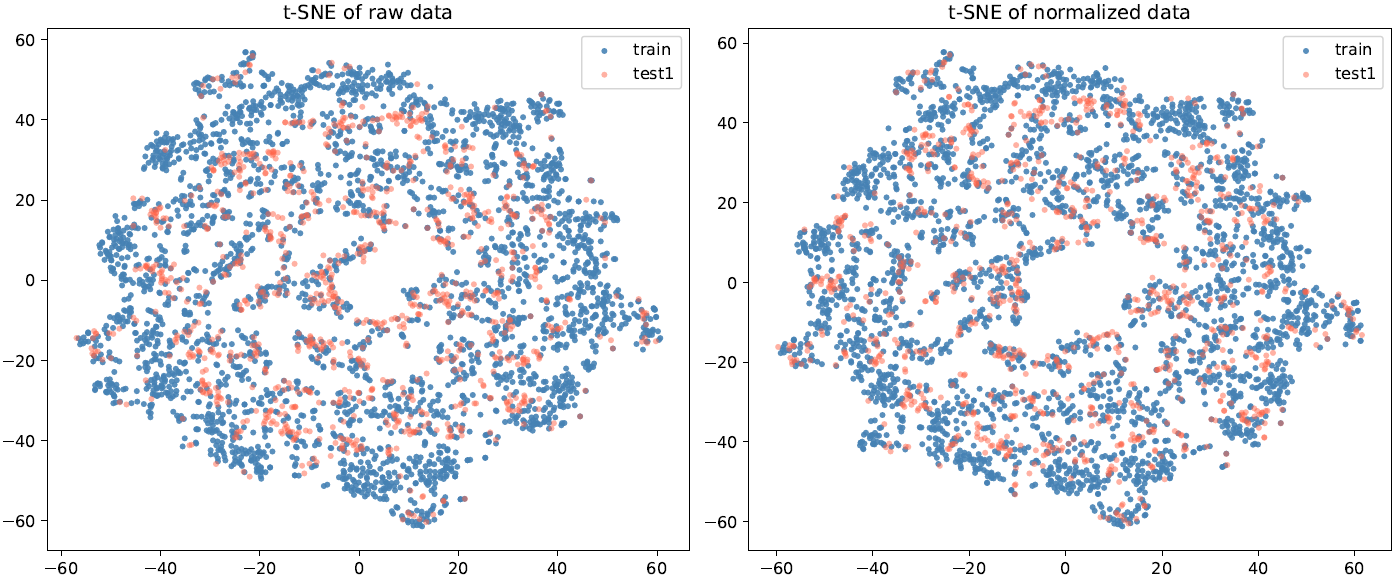}
    \caption{\textsc{Solar}, temporal setting}
    \label{fig:tsne_solar_temp}
\end{figure}

\begin{figure}[!h]
    \centering
    \includegraphics[width=0.6\linewidth]{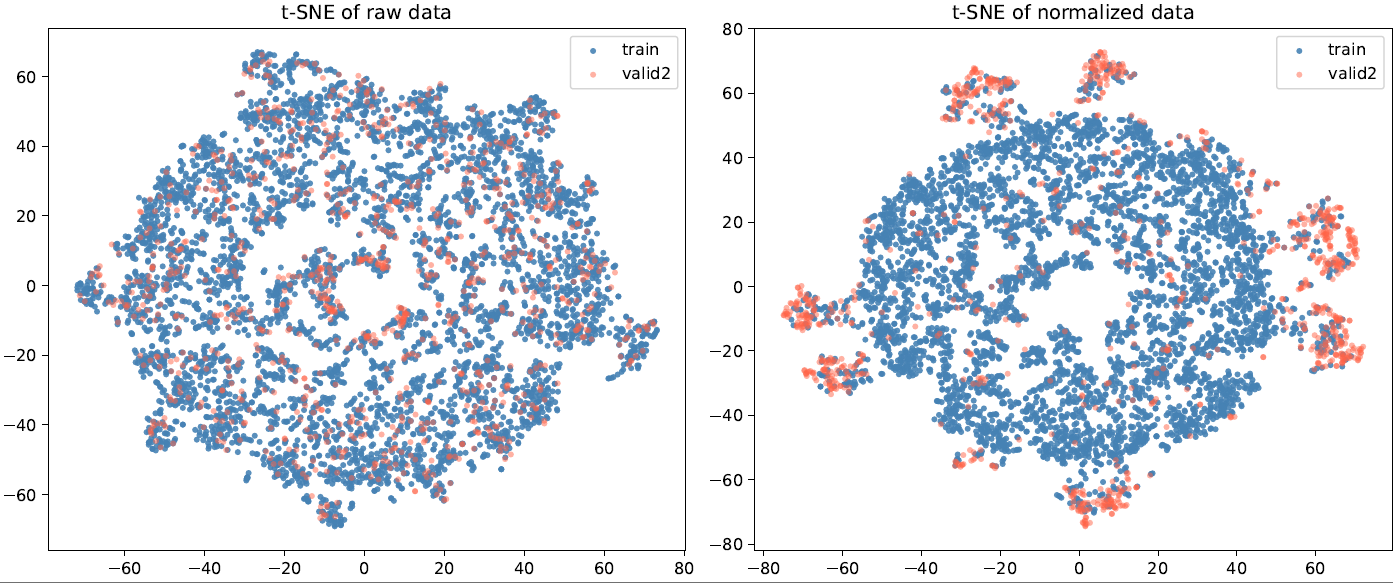}
    \caption{\textsc{Solar}, spatial setting}
    \label{fig:tsne_solar_spat}
\end{figure}

\begin{figure}[!h]
    \centering
    \includegraphics[width=0.6\linewidth]{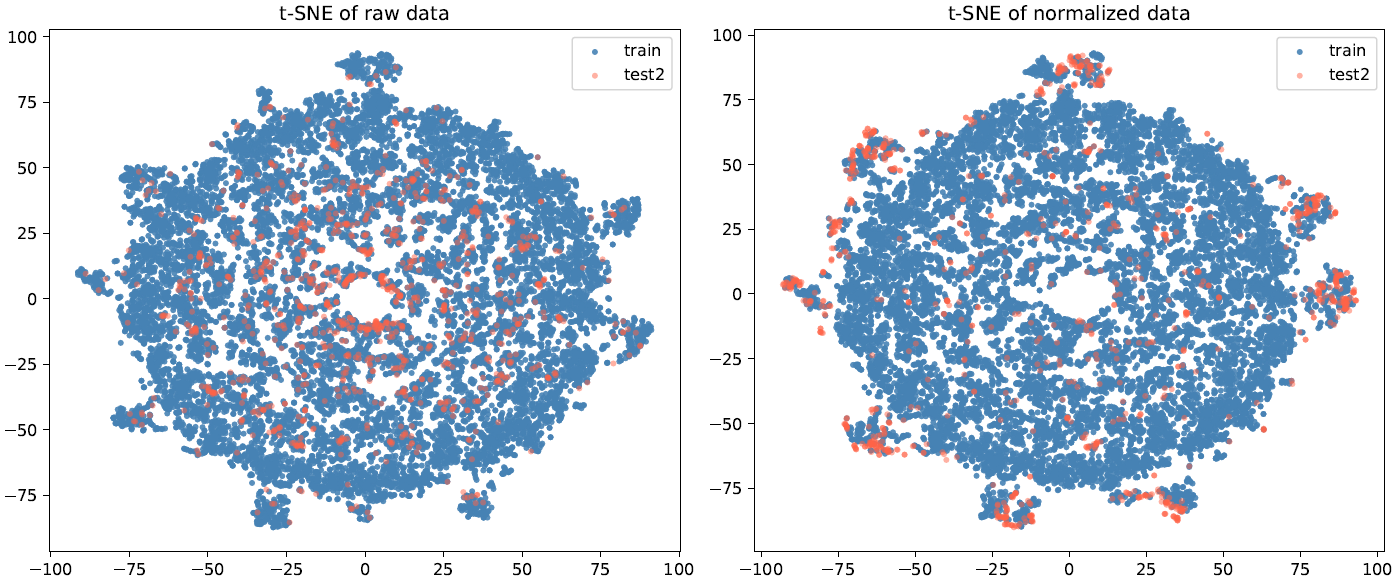}
    \caption{\textsc{Solar}, Overall setting}
    \label{fig:tsne_solar_cond}
\end{figure}

\begin{figure}[!h]
    \centering
    \includegraphics[width=0.6\linewidth]{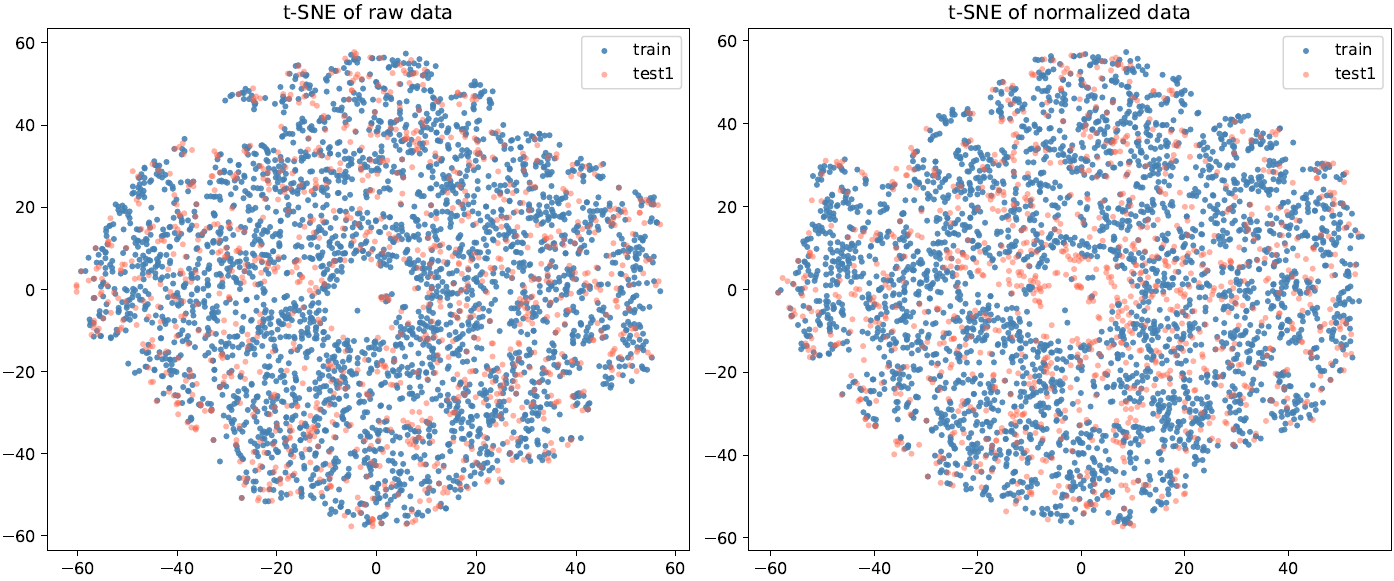}
    \caption{\textsc{Traffic}, temporal setting}
    \label{fig:traffic_time_tsne}
\end{figure}

\begin{figure}[!h]
    \centering
    \includegraphics[width=0.6\linewidth]{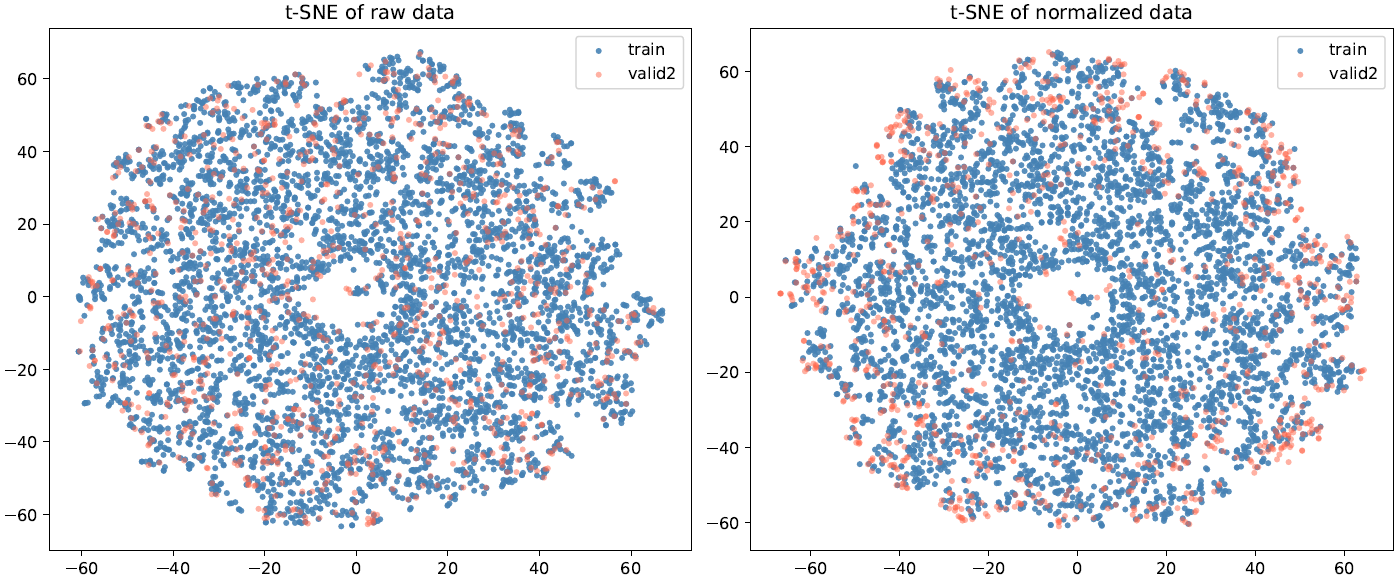}
    \caption{\textsc{Traffic}, spatial setting}
    \label{fig:traffic_indiv_tsne}
\end{figure}

\begin{figure}[!h]
    \centering
    \includegraphics[width=0.6\linewidth]{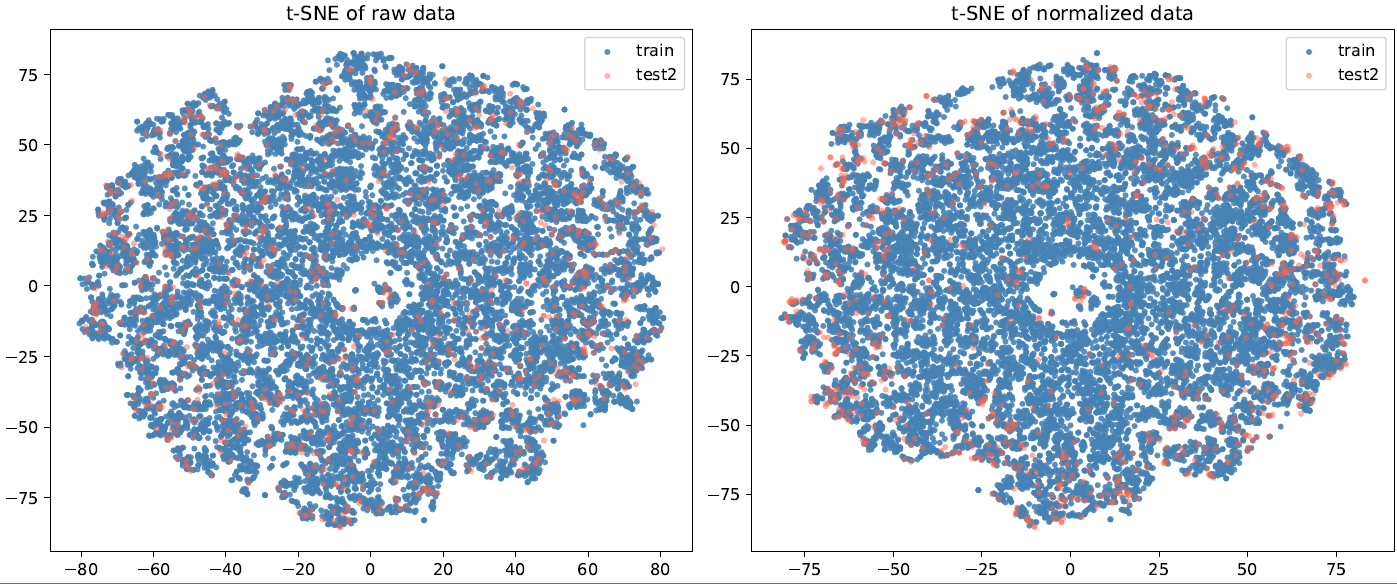}
    \caption{\textsc{Traffic}, Overall setting}
    \label{fig:traffic_cond_tsne}
\end{figure}

\begin{figure}[!h]
    \centering
    \includegraphics[width=0.6\linewidth]{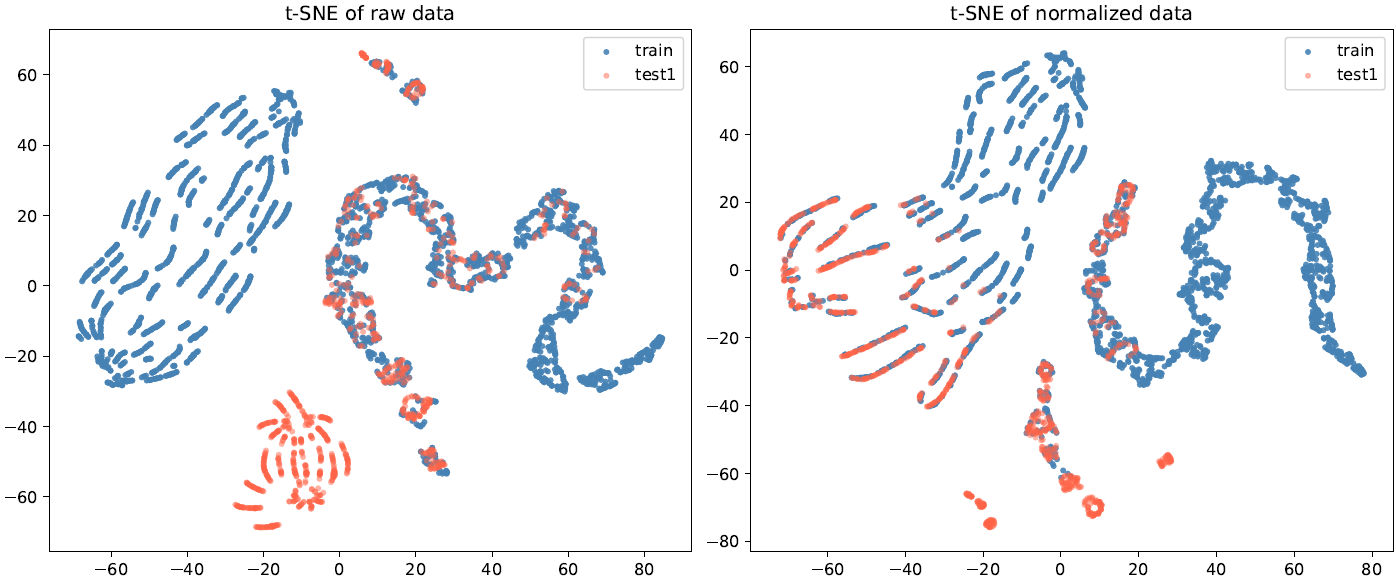}
    \caption{\textsc{Synthetic}, temporal setting}
    \label{fig:synthetic_time_tsne}
\end{figure}

\begin{figure}[!h]
    \centering
    \includegraphics[width=0.6\linewidth]{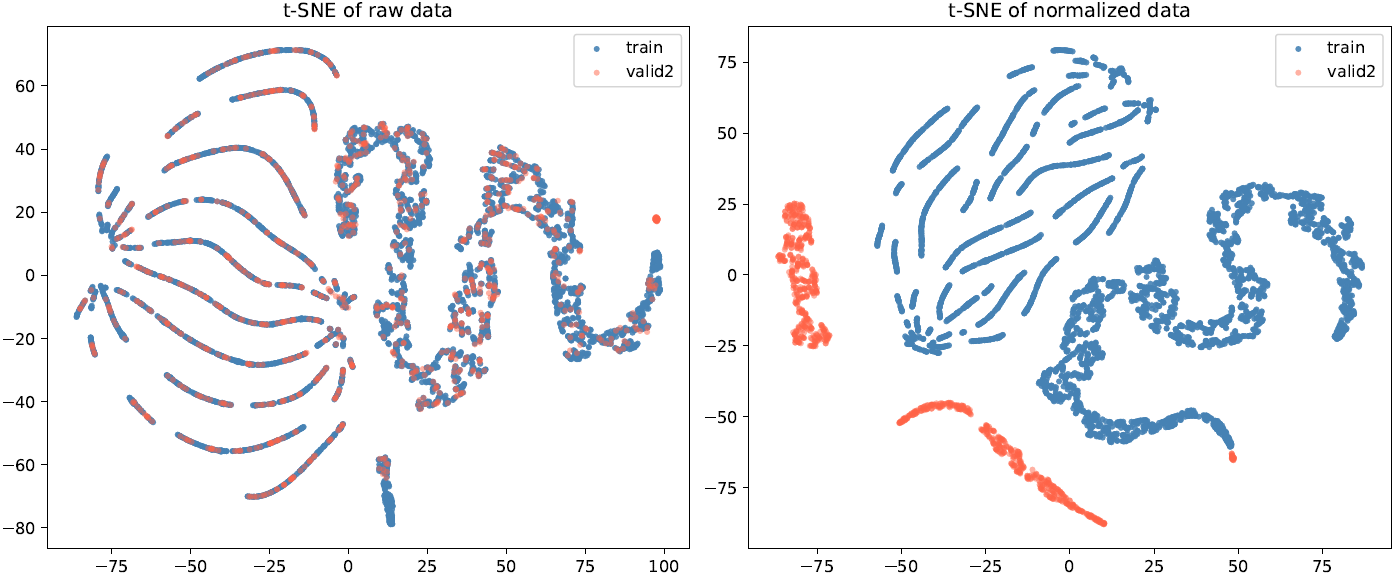}
    \caption{\textsc{Synthetic}, spatial setting}
    \label{fig:synthetic_indiv_tsne}
\end{figure}

\begin{figure}[!h]
    \centering
    \includegraphics[width=0.6\linewidth]{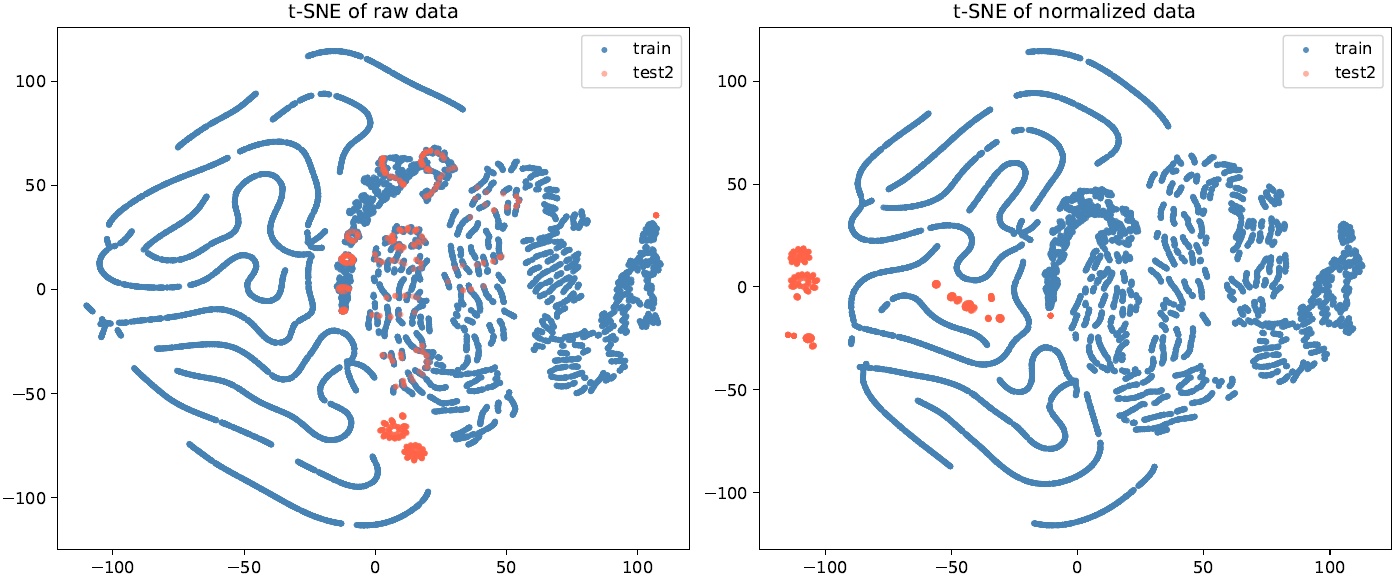}
    \caption{\textsc{Synthetic}, Overall setting}
    \label{fig:synthetic_cond_tsne}
\end{figure}

%% file: plots/distances.tex
\begin{table}[!h]
\centering
    \scalebox{0.6}{
    \begin{tabular}{lcccccc}
    \toprule
         & \multicolumn{2}{c}{None} & \multicolumn{2}{c}{Standard} & \multicolumn{2}{c}{Instance}\\
          \cmidrule(r){2-3}  \cmidrule(r){4-5}  \cmidrule(r){6-7}
        & \thead{Temporal} & \thead{Spatial} & \thead{Temporal} & \thead{Spatial} & \thead{Temporal} & \thead{Spatial} \\
    \midrule
        Electricity & 7.09 & 57.16 &\textbf{ 0.05} &\textbf{ 0.42} & 0.64 & 1.01 \\
        Solar       & 3.48 & 0.98 & 1.05 & 0.3 & \textbf{0.78} & \textbf{0.29} \\
        Traffic     & \textbf{0.32} & 0.99 & 1.39 & 4.29 & 0.92 & \textbf{0.45} \\
        Synthetic   & 15.97 & 39.51 & 2.53 & 6.26 & \textbf{0.9} & \textbf{3.76} \\
    \midrule
    Difference      & 0\% & 0\% & -20.27\% & -20.13\% & 18.86\% & 78.42\%\\
    \bottomrule
    \end{tabular}
    }
    \caption{Energy distances between shifted distributions (inputs).} 
    \label{tab:distances_energy_a}
\end{table}

\begin{table}[!h]
\centering
    \scalebox{0.6}{
    \begin{tabular}{lcccccc}
    \toprule
         & \multicolumn{2}{c}{None} & \multicolumn{2}{c}{Standard} & \multicolumn{2}{c}{Instance}\\
          \cmidrule(r){2-3}  \cmidrule(r){4-5}  \cmidrule(r){6-7}
        & \thead{Temporal} & \thead{Spatial} & \thead{Temporal} & \thead{Spatial} & \thead{Temporal} & \thead{Spatial} \\
    \midrule
        Electricity & 7.31 & 59.13 & \textbf{0.05} & \textbf{0.43} & 0.67 & 1.05 \\
        Solar       & 3.56 & 1.05  & 1.08 & \textbf{0.32} & \textbf{0.88} & \textbf{0.32} \\
        Traffic     & \textbf{0.32} & 1.02  & 1.4  & 4.43 & 0.93 & \textbf{0.47 }\\
        Synthetic   & 16.51 & 40.77 & 2.62 & 6.46 & \textbf{0.93} & \textbf{4.38} \\
    \midrule
    Difference      & 0\% & 0\% & -21.1\% & -20.34\% & 17.46\% & 77.73\%\\
    \bottomrule
    \end{tabular}
    }
    \caption{Energy distances between shifted distributions (windows).} 
    \label{tab:distances_energy_b}
\end{table}

\begin{table}[!h]
\centering
    \scalebox{0.6}{
    \begin{tabular}{lcccccc}
    \toprule
         & \multicolumn{2}{c}{None} & \multicolumn{2}{c}{Standard} & \multicolumn{2}{c}{Instance}\\
          \cmidrule(r){2-3}  \cmidrule(r){4-5}  \cmidrule(r){6-7}
        & \thead{Temporal} & \thead{Spatial} & \thead{Temporal} & \thead{Spatial} & \thead{Temporal} & \thead{Spatial} \\
    \midrule
        Electricity & 3.01 & 20.66 & 0.02 & 0.15 & \textbf{0} & \textbf{0} \\
        Solar       & 1.18 & 0.3   & 0.36 & 0.09 & \textbf{0} & \textbf{0} \\
        Traffic     & 0.08 & 0.4   & 0.36 & 1.72 & \textbf{0} & \textbf{0} \\
        Synthetic   & 4.51 & 10.99 & 0.71 & 1.74 & \textbf{0} & \textbf{0} \\
    \midrule
    Difference      & 0\% & 0\% & -24.23\% & -19.14\% & 100\% & 100\%\\
    \bottomrule
    \end{tabular}
    }
    \caption{Energy distances between shifted distributions (statistics).} 
    \label{tab:distances_energy_c}
\end{table}

\begin{table}[!h]
\centering
    \scalebox{0.6}{
    \begin{tabular}{lcccccc}
    \toprule
         & \multicolumn{2}{c}{None} & \multicolumn{2}{c}{Standard} & \multicolumn{2}{c}{Instance}\\
          \cmidrule(r){2-3}  \cmidrule(r){4-5}  \cmidrule(r){6-7}
        & \thead{Temporal} & \thead{Spatial} & \thead{Temporal} & \thead{Spatial} & \thead{Temporal} & \thead{Spatial} \\
    \midrule
        Electricity & 0.08 & 0.57 & \textbf{0.08} & 0.57 & \textbf{0.08} & 0.57 \\
        Solar       & 0.29 & \textbf{0.10}  & 0.29 & \textbf{0.10}  & 0.29 & \textbf{0.10} \\
        Traffic     & \textbf{0.16} & 0.17 & 0.17 & \textbf{0.15} & \textbf{0.16} & 0.17 \\
        Synthetic   & \textbf{0.10}  &\textbf{ 2.04} & \textbf{0.10}  &\textbf{ 2.04} & \textbf{0.10}  & \textbf{2.04} \\
    \midrule
    Difference      & 0\% & 0\% & -1.56\% & 2.94\% & 0\% & 0\%\\
    \bottomrule
    \end{tabular}
    }
    \caption{Energy distances between shifted distributions (modulations).} 
    \label{tab:distances_energy_d}
\end{table}